\documentclass[a4paper,twoside,openright,titlepage,
               headinclude,,footinclude,BCOR5mm,
               numbers=noenddot,cleardoublepage=empty,
               tablecaptionabove]{scrreprt}
               
\usepackage[T1]{fontenc}
\usepackage[utf8]{inputenc}
\usepackage[english]{babel}
\usepackage{amsmath,amssymb}
\usepackage{indentfirst}
\usepackage[style=philosophy-modern,hyperref, firstinits=true]{biblatex}
\usepackage{calc}
\usepackage{listings}
\usepackage{graphicx}
\usepackage{shapepar}
\usepackage{pifont}
\usepackage[eulerchapternumbers,beramono,eulermath,pdfspacing,listings]{classicthesis}
\usepackage{arsclassica}
\usepackage{csquotes}
\usepackage{float}
\usepackage{subcaption}                         
\usepackage{sidenotes}
\usepackage{algorithm}                          
\usepackage[noend]{algpseudocode}
\usepackage[object=vectorian]{pgfornament}      
\usepackage[draft]{todonotes}                   

\usepackage[alpha]{third_party/mdpn}

\renewcommand\formatchapter[1]{%
     \begin{minipage}[b]{0.15\linewidth}
     \chapterNumber
     \end{minipage}%
     \begin{minipage}[b]{0.70\linewidth}
     \raggedright\spacedallcaps{#1}
     \end{minipage}
}

\newcommand{\myName}{Andrei-Marian Manolache}
\newcommand{\myTitle}{Deep Anomaly Detection in Text}
\newcommand{\mySubTitle}{}
\newcommand{\mySupervisor}{Conf. Dr. Bogdan Alexe}
\newcommand{\myFaculty}{Faculty of Mathematics and Computer Science}
\newcommand{\myUniversity}{University of Bucharest}
\newcommand{\myDate}{June 2021}

\DeclareRobustCommand*{\classicthesis}{Classic\-Thesis}
\DeclareRobustCommand*{\arsclassica}{{\normalfont\sffamily ArsClassica}}

\usepackage{bm}
\usepackage{booktabs}
\usepackage{xcolor}
\usepackage{mathtools}
\usepackage{amsmath}
\usepackage{amssymb}
\usepackage{multirow}
\usepackage[toc,page]{appendix}

\newcommand{\bx}{\bm{x}}

\newcommand{\E}{\mathop{\mathbb{E}}}

\usepackage{bm}

\newcommand{\Loss}[2][]{\mathcal{L}_{#1}\ifthenelse{\equal{#2}{}}{}{(#2)}}

\DeclarePairedDelimiterX{\infdivx}[2]{[}{]}{%
  #1\;\delimsize\|\;#2%
}

\DeclarePairedDelimiterX{\condx}[2]{[}{]}{%
  #1\mid#2%
}

\DeclarePairedDelimiterX{\br}[1]{[}{]}{#1}
\newcommand{\Expected}[1][]{
   \ifthenelse{\equal{#1}{}}{\mathbb{E}\br}{\mathbb{E}_{#1}\br}%
}
\newcommand{\Entropy}[1][]{
   \ifthenelse{\equal{#1}{}}{\mathbb{H}\br}{\mathbb{H}_{#1}\br}%
}

\newcommand{\keyword}[1]{\marginnote{\footnotesize{\textit{#1}}}}

\newcommand{\mail}[1]{\href{mailto:#1}{\texttt{#1}}}

\graphicspath{{graphics/}}

\usepackage{tabularx} 
\definecolor{lightergray}{gray}{0.99}

\newcommand{\meta}[1]{$\langle${\normalfont\itshape#1}$\rangle$}

\lstnewenvironment{code}%
   {\setkeys{lst}{columns=fullflexible,keepspaces=true}%
   \lstset{basicstyle=\small\ttfamily}}{}

\bibliography{bibliography}

\defbibheading{bibliography}{%
\cleardoublepage
\manualmark
\phantomsection
\addcontentsline{toc}{chapter}{\tocEntry{\bibname}}
\chapter*{\bibname\markboth{\spacedlowsmallcaps{\bibname}}
{\spacedlowsmallcaps{\bibname}}}}

\begin{document}
\pagenumbering{roman}
\pagestyle{plain}

\begin{titlepage}
\pdfbookmark{Titlepage}{Titlepage}
\changetext{}{140pt}{}{((\paperwidth  - \textwidth) / 2) - \oddsidemargin - \hoffset - 1in}{}
    \begin{center}

        {\LARGE  
            \hfill
            \vfill

            {\color{Maroon}\spacedallcaps{Deep Anomaly Detection in Text}}
            \\
            \bigskip
        }

        \vfill
        \pgfornament{93} \\
        \bigskip
		\mySubTitle
        \vfill

        \myName \\
        \bigskip
        \textit{coordinated by} \\
        \mySupervisor
        \bigskip

        \myFaculty \\
        \myUniversity

        \vfill                      

    \end{center}        
\end{titlepage} 

\thispagestyle{empty}
\pdfbookmark{Titleback}{Titleback}

\hfill

\vspace{\stretch{1}}

\noindent
\myName: \textit{\myTitle} MSc thesis, \myDate. \\


\noindent
{\raisebox{0.5ex}{\pgfornament[scale=0.2]{31}}\,\mail{andrei\_mano@outlook.com}

\bigskip
\noindent
This MSc thesis was written with \LaTeX{} using \arsclassica, a reworking of
the \classicthesis{} style designed by Andr\'e Miede, inspired to the
masterpiece \emph{The Elements of Typographic Style} by Robert Bringhurst. This template was provided by Florin-Radu Gogianu.}

\clearpage


\pdfbookmark[1]{Abstract}{Abstract} 

\begingroup
\let\clearpage\relax
\let\cleardoublepage\relax
\let\cleardoublepage\relax

\chapter*{Abstract} 

\noindent
Deep anomaly detection \keyword{Anomaly Detection (AD)} methods have become increasingly popular in recent years, with methods like Stacked Autoencoders \keyword{Autoencoder (AE)}, Variational Autoencoders and Generative Adversarial Networks \keyword{Generative Adversarial Network (GAN)} greatly improving the state-of-the-art. 
Other methods rely on augmenting classical models (such as the One-Class Support Vector Machine\keyword{One-Class Support Vector Machine (OC-SVM)}), by learning an appropriate kernel function using Neural Networks.
Recent developments in representation learning by self-supervision\keyword{Neural Network (NN)} are proving to be very beneficial in the context of anomaly detection.
Inspired by the advancements in anomaly detection using self-supervised learning\keyword{Self-Supervised Learning (SSL)} in the field of computer vision, this thesis aims at developing a method for detecting anomalies by exploiting ``pretext tasks'' tailored for text corpora\keyword{Computer Vision (CV)}.
This approach greatly improves the state-of-the-art on two datasets - 20Newsgroups and AG News, for both semi-supervised and unsupervised anomaly detection, thus proving the potential\keyword{Natural Language Processing (NLP)} for self-supervised anomaly detectors in the field of natural language processing.
\endgroup			

\vfill

\clearpage

\begin{flushright}
\itshape
The truth is rarely pure and never simple. \medskip
--- Oscar Wilde
\end{flushright}

\pdfbookmark{Acknowledgements}{Acknowledgements}

\chapter*{Acknowledgements}
\noindent
This thesis is based on the research performed by the author over the last years at Bitdefender's Theoretical Research laboratory. I am extremely grateful for the fruitful discussions I've had with my laboratory colleagues and for the guidance from various professors at the University of Bucharest during my Master's studies.

\noindent
In particular, I would like to thank to my thesis advisor, professor Bogdan Alexe, for the provided supervision, counseling and for the excellent courses held at the Faculty of Mathematics and Computer Sciences.

\noindent
The thesis is based on the paper \autocite{DATE}, therefore I would like to thank my co-authors Florin Brad and Elena Burceanu for their contributions, support, and inexhaustible patience.

\noindent
I would also like to thank to professor Marius Popescu for his insights and advises on the problem of authorship detection, and to Sergiu Nisioi for the suggestions regarding further developments.

\noindent
Parts of this work have been supported by UEFISCDI, under Project PN-III-P2-2.1-PTE-2019-0532.
\pagestyle{scrheadings} 

\phantomsection
\pdfbookmark{\contentsname}{tableofcontents}
\setcounter{tocdepth}{2}
\tableofcontents
\markboth{\spacedlowsmallcaps{\contentsname}}{\spacedlowsmallcaps{\contentsname}} 
\cleardoublepage
\pagenumbering{arabic}


\chapter{Prologue}
\label{chp:prologue}

\section{Introduction}
\label{sec:introduction}
\noindent
\paragraph{Anomaly detection} is the task of detecting data points that are deviating from the expected data distribution (inliers). Such a data point\keyword{inlier, anomaly, outlier, abnormality, deviant} is said to be an \textit{outlier}, \textit{abnormality}, or \textit{deviant}. Hawkins \autocite{Hawkins1980} notably defined an anomaly as follows:

\begin{displayquote}
``\textit{An outlier is an observation which deviates so much from the other observations as to arouse suspicions that it was generated by a different mechanism}.''
\end{displayquote}

 The early work on anomaly detection was historically produced in the statistics community, but the rapid growth of available data banks and the development of computer hardware and software solutions has led to increasing amounts of interest from Data Scientists and the broader Computer Science community. 

The interest of researchers in outliers span several decades \autocite{KnorrNg,Chandola2009,Aggarwal2013}, some of the first formal mentions about "discordant observations"\keyword{discordant} going back to the $19^{th}$ century \autocite{EdgeworthXLIOD}. Efforts for developing Anomaly Detection methods have proven very fruitful, with applications in credit card fraud detection \autocite{Dorronsoro1997}, network monitoring \autocite{ids2017,rnn_network}, intrusion detection systems \autocite{Banoth2017}, time series \autocite{ts}, medical imaging \autocite{medical_anomaly}, and manufacturing \autocite{Kammerer2019}.

\begin{figure}[h!]
    \centering
    \includegraphics[width=1\textwidth]{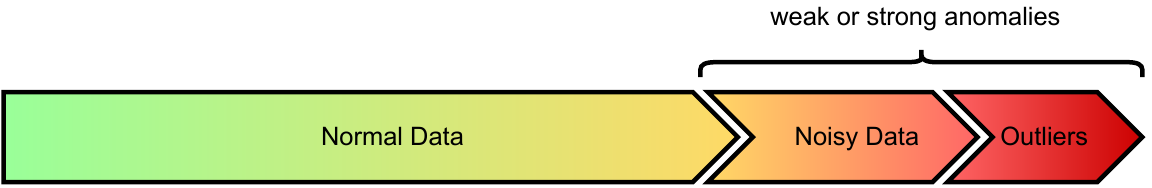}
    \caption{The anomaly spectrum. Noisy data can be regarded as ``weakly'' anomalous. (figure adapted from: \cite{Aggarwal2013})}
    \label{fig:outlier-spectrum}
\end{figure}

Most anomaly detection systems are usually producing an \textit{anomaly score} \keyword{anomaly score} for a given data sample. Such a score is a very general form of output and can be used to asset the ``abnormality degree'' of the sample (see Fig. \ref{fig:outlier-spectrum}). Such a scoring method is desirable since an anomaly highly depends on the intended use-case and the tolerance to false-positives that the system should have. For example, we would not want an automatic fraud detector used in a banking system to flag unusual (but otherwise benign) transactions as being fraudulent. Having a granular scoring scheme enables us to possibly forward such a transaction to further evaluation by some human in the loop.

\section{Contribution}
\label{sec:contribution}
\noindent
In this thesis, we present various architectures for anomaly detection in text. Particularly, we introduce a novel deep learning method based on the Transformer architecture, called DATE\keyword{Transformer, DATE} \autocite{DATE}. To our knowledge, DATE is the first end-to-end deep anomaly detection method on text that is using self-supervision to produce an anomaly score. Our contributions are summarized below:

\begin{itemize}
    \item We introduce a sequence-level self-supervision task called \emph{Replaced Mask Detection} \keyword{Replaced Mask Detection (RMD), Replaced Token Detection (RTD)} that, when used in conjunction with the Replaced Token Detection task \autocite{Clark2020}, improves the anomaly detection capabilities of our model and stabilizes the training. RMD is used to distinguish between text that is corrupted in different ways.
    \item We formulate an efficient Pseudo Label \keyword{Pseudo Label (PL)} score for text anomalies. We remove the need of doing inference on text corrupted in multiple ways \autocite{e3outlier} by looking at the individual tokens probabilities. The token-level PL score significantly speeds up inference times and makes the results interpretable.
    \item We outperform existing state-of-the-art semi-supervised text AD models on two datasets: 20Newsgroups (+4.7\% AUROC) and AG News (+6.9\% AUROC). More so, when doing experiments in the unsupervised AD settings, DATE surpasses all other methods trained with 0\% contamination, even at 10\% train data contamination rate.
    \item We prove that anomaly detectors in text can be used for authorship detection by training the models on the individual authors' text corpus and treating other texts as anomalies.
\end{itemize}

\section{Outline}
\label{sec:outline}

\noindent
The rest of this thesis is organized as follows:

Chapter \ref{chp:rl} introduces notions about representation learning \keyword{representation learning}, with an emphasis on deep self-supervised learning \keyword{self-supervised learning} for text. We briefly discuss about models that are able to produce neural word embeddings \keyword{word embedding}, then we focus on some neural architectures that can be used to deal with sequential data. We study two particular neural language models based on the Transformer\keyword{Transformer} \autocite{Vaswani2017} architecture.

In Chapter \ref{chp:anomalydetection} we introduce the problem of anomaly detection, which is the main concern of this thesis. We present some classical models and the recent advancements that are using deep neural networks.

We finally introduce our solution for detecting anomalies in text using deep learning in Chapter \ref{chp:date}. We describe how we exploit our ``pretext task'' \keyword{pretext tasks} to do training and inference, then we go in detail about our experimental setup, quantitative and qualitative results, and a rigorous ablation study. We also showcase an application for anomaly detection in text, in the form of authorship detection in the supplementary material \ref{apx:authorship}.


\chapter{Representation Learning}
\label{chp:rl}
\begin{figure}[!b]
    \centering
    \includegraphics[width=1\textwidth]{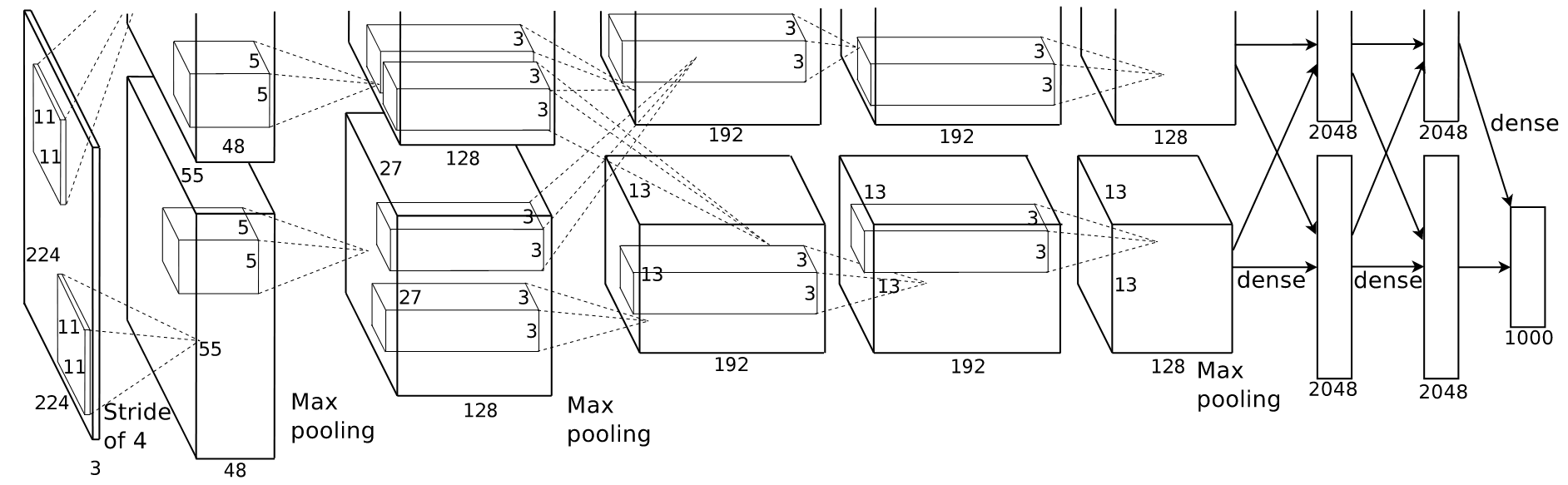}
    \caption{The AlexNet CNN architecture. (figure from: \cite{alexnet})}
    \label{fig:alexnet}
\end{figure}

\noindent
\paragraph{Neural networks} have recently seen great popularity, mainly due to the fact that AI systems based on neural networks are able to learn from massive amounts of data, given enough compute power. The development of GPU acceleration techniques \autocite{gpunn} and dedicated hardware accelerators \autocite{tpu} has enabled researchers and practitioners to train neural networks that can achieve human-level or even super-human performance on various tasks \autocite{CiresanTraffic, AtariSuperhuman}. The popularity of supervised deep learning approaches has escalated after the 2012 edition of the \textit{ImageNet Large Scale Visual Recognition Challenge} \autocite{imagenet}, where Alex Krizhevsky has obtained a top-5 error of $15.3\%$ with AlexNet, a convolutional neural network \keyword{Convolutional Neural Network (CNN)} \autocite{fukushima, waibel, lecun_cnn, lecun_cnn2}. This approach brought an improvement of $41.4\%$ over the next best solution \autocite{alexnet}, and was the start of a complete shift in interest from manually constructed features to learned features. The AlexNet architecture is shown in Fig. \ref{fig:alexnet}.

This renewed interest in deep neural networks has led to many different architectures which greatly improved the state-of-the-art in computer vision \autocite{googlenet, vgg, resnet}, natural language processing \autocite{w2v, Hochreiter1997, gru, Vaswani2017}, and deep reinforcement learning \autocite{atari, dota2, starcraft, go}.

Most of the architectures used in computer vision were trained via supervised learning\keyword{supervised learning}, where models are instructed to solve particular tasks. The idea of transfer learning \keyword{transfer learning} \autocite{transfer_nips} by using some of the learned weights and extracting representations \keyword{representations, fine-tuning} or fine-tuning the model quickly re-emerged and is still widely used for training when limited amounts of data are provided \autocite{transfer_survey}. In natural language processing, since text is unstructured and varies in length, it was already very desirable to learn general word representations which could be used in conjunction with specialized models. This has led to the development of many techniques for representation learning in NLP \autocite{w2v, glove, fasttext} culminating with deep language\keyword{deep language models} models \autocite{elmo} based on the Transformer architecture \autocite{Vaswani2017, bert, gpt3, Clark2020}, which are usually trained via self-supervision\keyword{self-supervision}.

\begin{figure}[!t]
    \centering
    \includegraphics[width=1\textwidth]{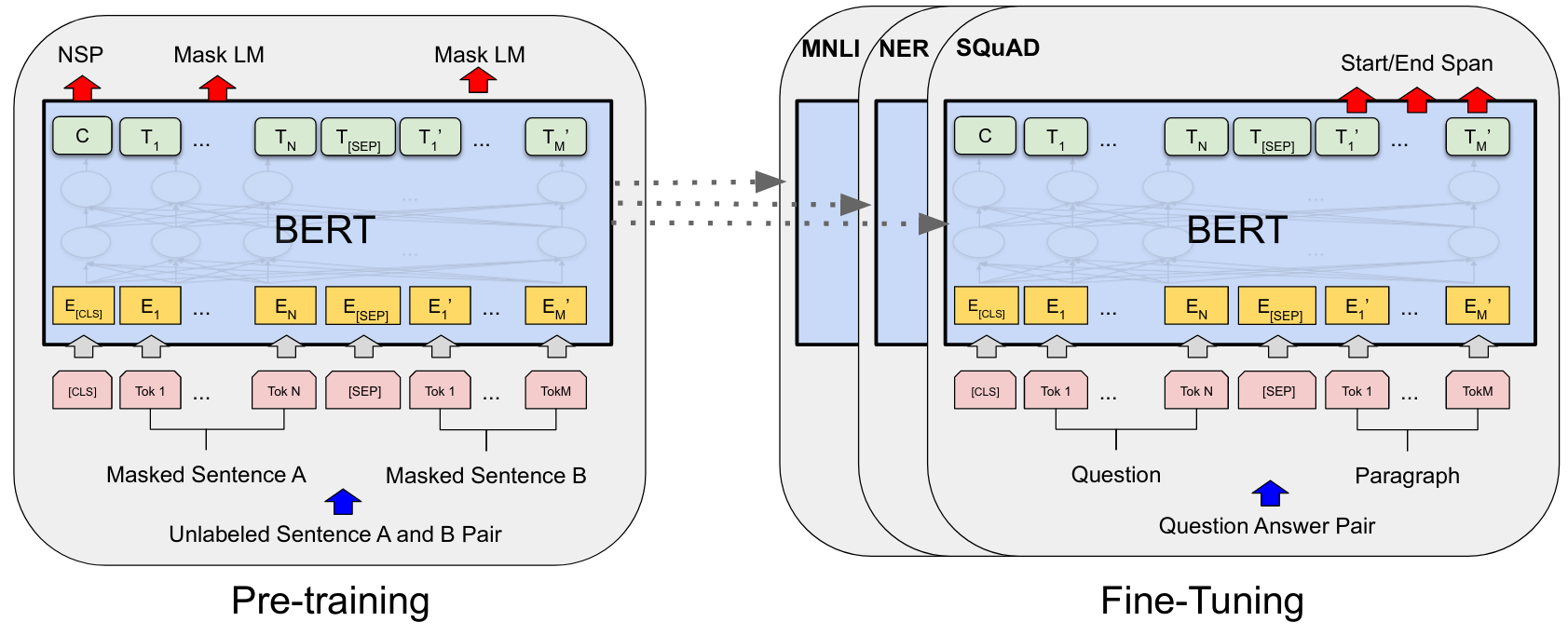}
    \caption{The training and fine-tuning procedures for the BERT Transformer-based model. More details about Transformer models will be provided in section \ref{sec:transformer}. The BERT architecture will be detailed in subsection \ref{sub:bert}. (figure from: \cite{bert})}
    \label{fig:bert}
\end{figure}

\section{Self-Supervised Learning}
\label{sec:ssl}
\noindent
As previously mentioned, much of the recent progress in the field of machine learning is due to models trained to solve particular tasks in a supervised fashion. These models require large collections of (usually) expensive labeled data. One way of alleviating the requisite of annotated data is by transferring knowledge from models trained on immense labeled data banks. This approach assumes that these kinds of data banks exist, which can lead to domain-specific solutions or, at best, sub-optimal results. For instance, the knowledge transfer from a 2D computer vision dataset to a 3D task can be quite limited. Similarly, we can't expect that the weights of a model trained on pictures of cats and dogs to be sufficient for classifying medical images. Moreover, the task of labeling everything in the world is impossible. These limitations can make us wonder if there's a more efficient way of learning general representations. One way of tackling this problem is self-supervised learning.

Self-supervised learning is a form of unsupervised learning where the data provides the supervision \autocite{az_inria}. Generally, the self-supervised approach is done by corrupting a part of the input and making the model reconstruct some of the lost information. In this way, the model should be able to learn the general structure of the data in an unsupervised way, thus enabling us to use the learned representations for other downstream tasks\keyword{downstream task} via transfer learning. The self-supervision task is sometimes referred to as a ``\textit{pretext task}'', and the loss function is usually called ``\textit{proxy loss}''\keyword{proxy loss}. By using self-supervised learning, the model obtains supervisory signals from the structure of the data, therefore the reliance on labeled datasets is eliminated.

The field of natural language processing is the earliest success story of SSL. By using huge amounts of unlabeled text, deep language models trained using self-supervision resulted in which many consider being ``\textit{NLP's ImageNet moment}''. These models are usually given an input phrase that has some of its parts masked, and then they are assigned to complete the phrase. For example, when given the text ``\textit{The students are \textbf{(blank)} mathematics at the faculty of \textbf{(blank)}}'', the model should be able to determine that the most likely action for the students to do in this context is to learn and that the most common faculty where one learns mathematics is the faculty of mathematics, thus one viable (but not unique) completion would be ``\textit{The students are \textbf{learning} mathematics at the faculty of \textbf{mathematics}}''. As a result, these masked language models\keyword{Masked Language Model (MLM)} learn to represent the contexts and the significance of the words in different contexts \autocite{lecun_ssl_fb}.

\begin{figure}[!t]
    \centering
    \includegraphics[width=1\textwidth]{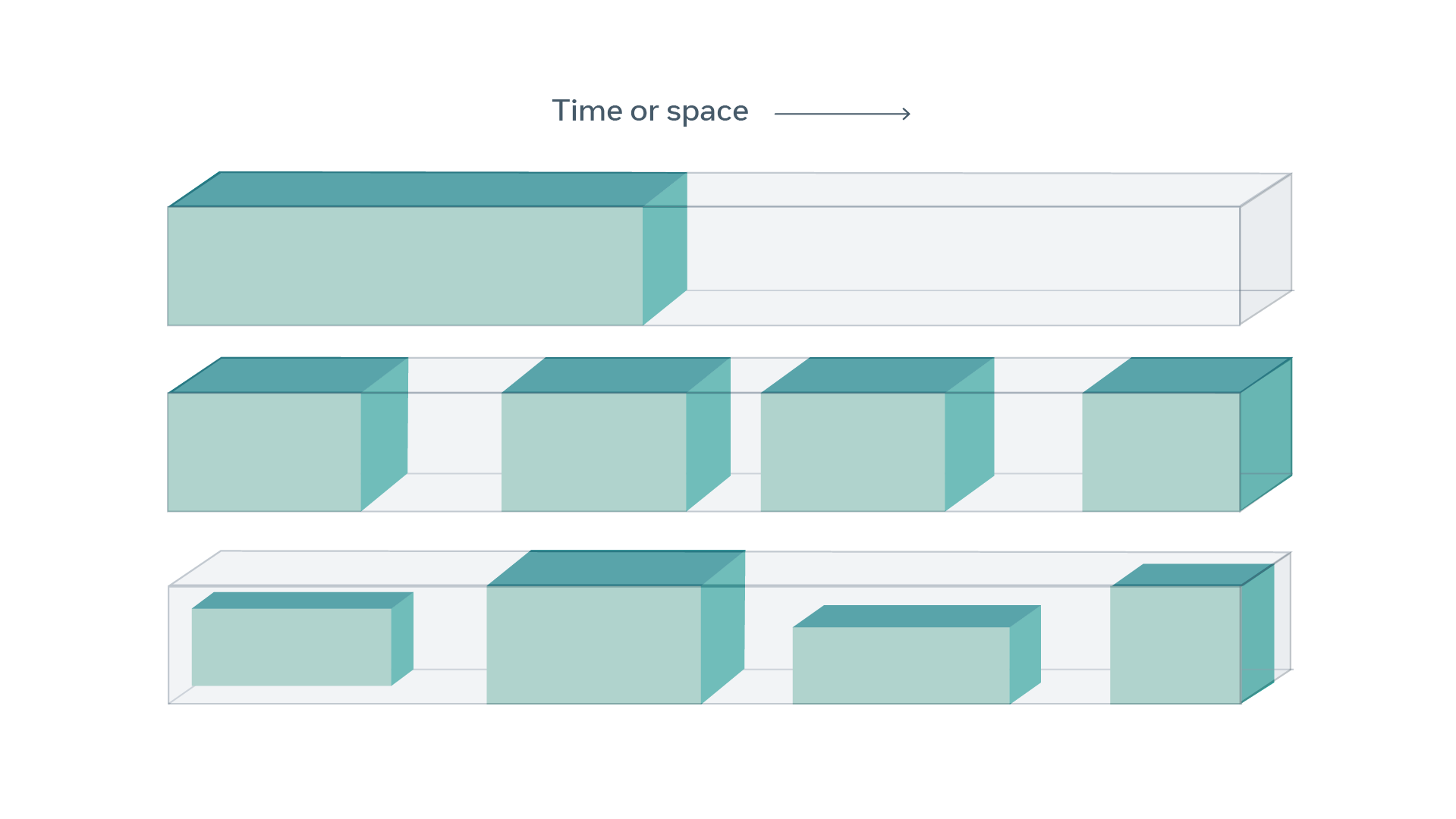}
    \caption{Example of general pretext tasks for self-supervised learning. In natural language processing we could try to predict the next token, or some masked tokens, from a sequence of tokens. In computer vision we can build a model that reconstruct the missing parts of a corrupted image. (figure from: \cite{lecun_ssl_fb})}
    \label{fig:ssl_lecun_fb}
\end{figure}

The development of SSL models is not limited to the field of NLP. Recent models trained with self-supervision tasks in the field of computer vision have obtained state-of-the-art results when fine-tuned to solve classification tasks \autocite{simclr1, byol}. Furthermore, some models trained via self-supervision and fine-tuned on fractions of the dataset are able to surpass their fully-supervised counterparts, even when using just $10\%$ labeled data \autocite{simclr2}. 

But still, CV has not benefited from SSL to the same extent as NLP, in part because in vision it's more difficult to represent uncertainty due to the domain in which we need to make predictions being virtually infinite. We are able to fix a finite vocabulary and get a probability distribution over the entire vocabulary for a masked word, but if given a picture of a book on a table, it's practically impossible to get a distribution over all the possible objects that may be on that table. As a result, SSL in vision is usually done by predicting various transformations (such as rotations or translations) of the input \autocite{e3outlier}, agreeing upon various views of the same object \autocite{byol}, or by using \textit{contrastive learning}\keyword{contrastive learning} \autocite{simclr1}.

In the next sections, we'll focus on self-supervised learning for natural language processing and we'll discuss some of the more common models and approaches.

\section{Word Embeddings}\label{sec:word-embeddings}  
\noindent
Word embeddings are dense vectorial representations of words. A key propriety of these embeddings is that the words that have similar semantic meaning also tend to have similar encodings.


The idea of obtaining dense vectors from words using neural networks is not new \autocite{bengio_lm}, but the \textit{Collobert-Weston model} \autocite{collobert-weston} was one of the first successful attempts at learning general-purpose word embeddings for multi-task learning. The authors describe a convolutional neural network which is leveraging unlabeled data using a language model that minimizes the following ranking-type loss:

\begin{equation}
    \sum_{s\in \mathcal{S}}\sum_{w\in \mathcal{D}}max(0, 1-f(s)+f(s^w))
\end{equation}

where $\mathcal{S}$ is the set of context windows of the text, $\mathcal{D}$ is the dictionary of words, and $f(\cdot)$ is the neural network without the softmax\sidenote{\textnormal{The softmax function, \textit{$\sigma(z)_i=\frac{e^{z_i}}{\sum_{j=1}^{K}e^{z_j}}$\\$\forall i\in\{1, \dots, K\}$ and $z=(z_1,\dots,z_K)$} is an activation function used to normalize the output of a network to a probability distribution over the predicted classes.}} layer, and $s^w$ is a context window where the middle word has been replaced by a random word $w$. In this way, the language model is trained to solve a binary classification task: whether the word in the middle of the input is related to the context or not.

This approach has led to improvements when sharing weights and jointly training the LM along with supervised tasks such as \textit{part-of-speech tagging}, \textit{named entity recognition}, \textit{chunking}, and \textit{semantic role labeling}. Furthermore, the model produced high-quality word embeddings as a side-effect (Table \ref{fig:emb-cw}).

\begin{table}[!b]
    \centering
    \includegraphics[width=1\textwidth]{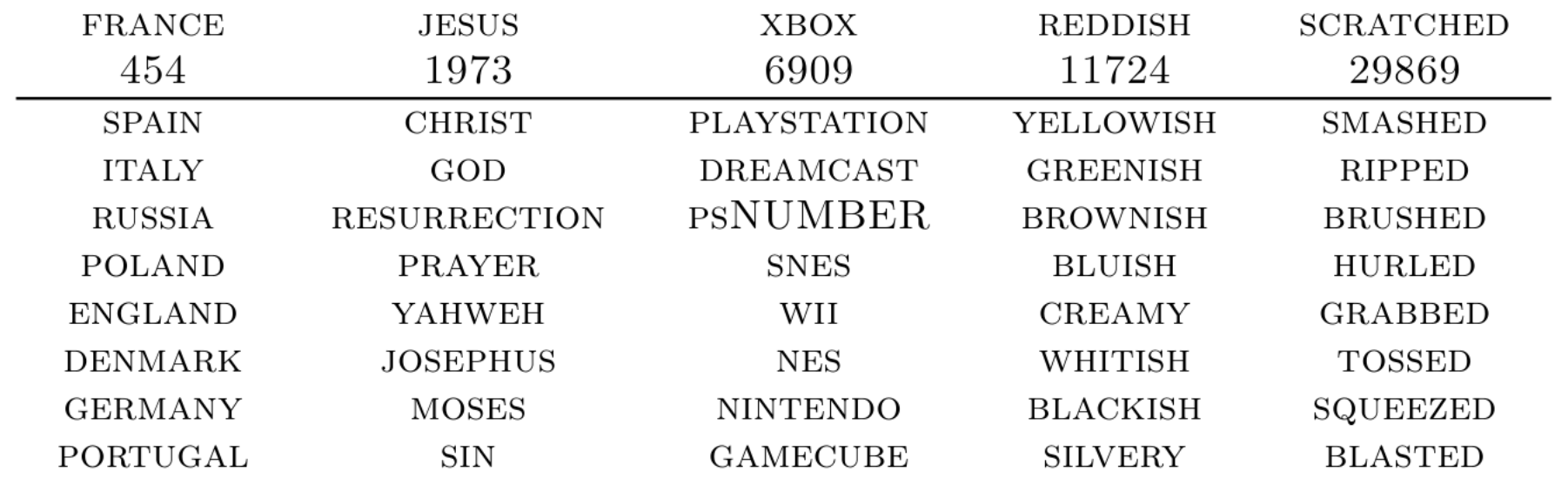}
    \caption{Qualitative result regarding the word embeddings produced by the Collobert-Weston model. The queried words are followed by their index in the dictionary and their nearest 8 neighbors. (table from: \cite{collobert-weston})}
    \label{fig:emb-cw}
\end{table}

\begin{figure}[!t]
    \centering
    \includegraphics[width=1\textwidth]{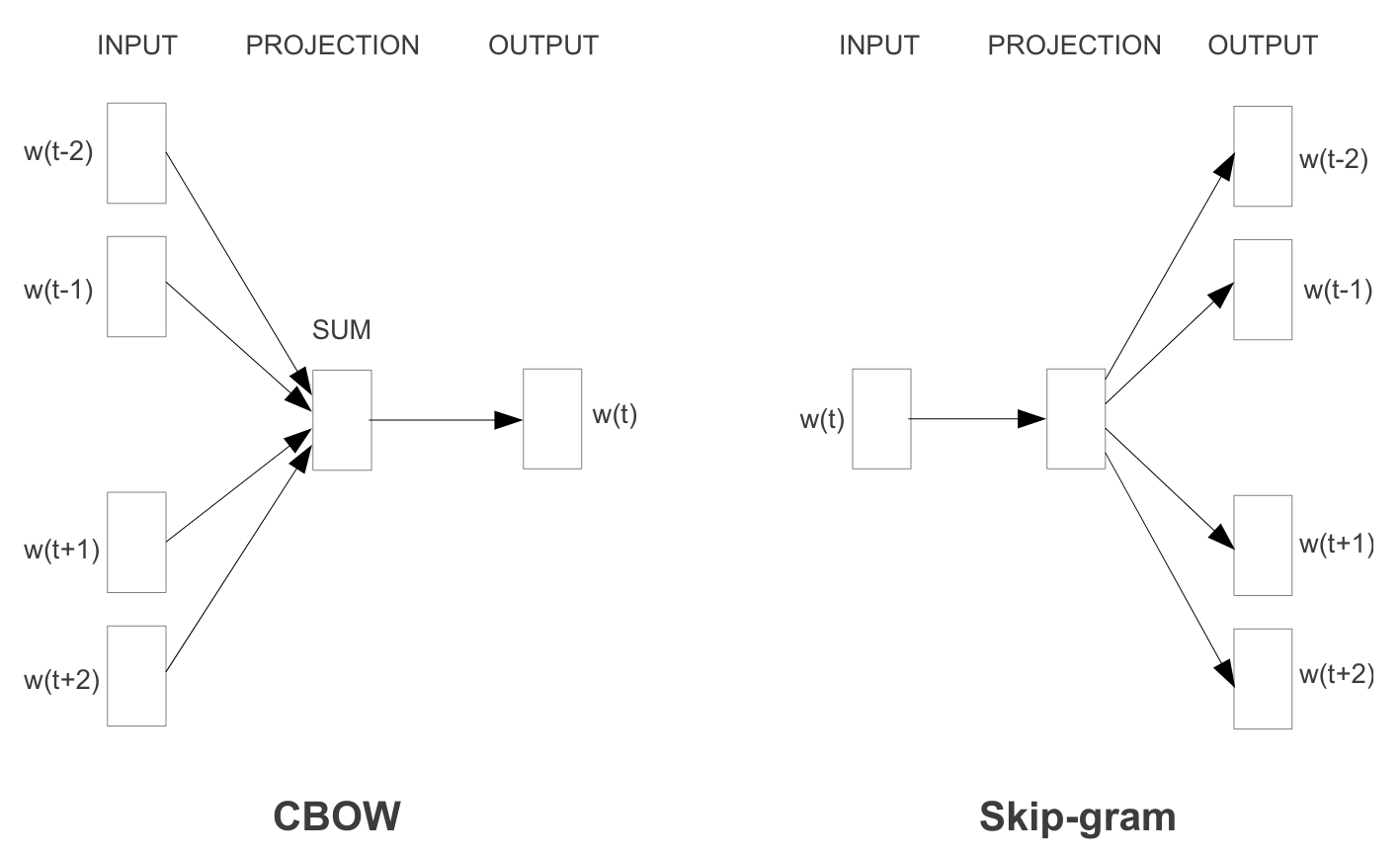}
    \caption{The CBOW and Skip-Gram models. CBOW is predicting the word with respect to the context, while Skip-Gram is predicting the context with respect to the word. (figure from: \cite{mikolov_cbow_skipgram})}
    \label{fig:cbow-skipgram}
\end{figure}

The next great leap for word embeddings was brought by a class of techniques called \textit{Word2Vec}\keyword{Word2Vec} \autocite{w2v, w2v2, mikolov_cbow_skipgram}. This time, the models are specifically trained to produce word embeddings, the quality of the embeddings being measured with a word similarity task and compared to the previous state-of-the-art. We'll briefly examine two of the models from the Word2Vec family: the \textit{Skip-Gram} model and the \textit{Continuous Bag of Words}\keyword{Skip-Gram, Continuous Bag of Words (CBOW)} (Fig. \ref{fig:cbow-skipgram}).

The Skip-gram model is using the Negative sampling objective to predict the context of a word, on the other hand, CBOW's objective is to\keyword{Negative sampling (NEG)} predict the word in the middle of the context. One key advantage of these models is that they require relatively low compute power, thus they can be trained on large corpora. The authors note that Skip-gram delivers better performance on small monolingual datasets, while CBOW is faster and benefits from large amounts of data.

Given a middle token $w_t$ and a training context of $k$ tokens, the Skip-gram model tries to compute the probability of the neighbouring tokens as being placed correctly. Formally, Skip-gram maximizes the following expression:

\begin{equation}
    \frac{1}{T}\sum_{t=1}^T[\sum_{j=-k}^k log \mathbb{P}(w_{t+j}|w_t)],
\end{equation}

\noindent
$\mathbb{P}(w_i|w_j)$ is defined as:

\begin{equation}
    \mathbb{P}(w_i|w_j) = \frac{e^{u_{w_i}^T v_{w_j}}}{ \sum_{\xi=1}^{|\mathcal{D}|} e^{ u_\xi^T v_{w_j} } },    
\end{equation}

\noindent
where $|\mathcal{D}|$ is the cardinality of the dictionary, $u_w$ is the associated input representation, and $v_w$ is the associated output representation.

The procedure for CBOW is analogous, except for the fact that we're now trying to maximize the log probability $log\mathbb{P}(w_t|w_{t-j} w_{t+j})$.

An interesting property of the learned vector representations is that linear operations seem to be able to capture semantic meaning. For example, the operation $vector("King") - vector("Man") + vector("Woman")$ yields a vector close to $vector("Queen")$ \autocite{mikolov-etal-2013-linguistic}, and $vector("France")-vector("Paris")$ is close to $vector("Italy") - vector("Rome")$ \autocite{mikolov_cbow_skipgram}.

In the following years, other popular methods based on the Skip-Gram and CBOW methods were developed, the most notable being \textit{GloVe}\keyword{Global Vectors (GloVe) fastText} \autocite{glove} and \textit{fastText} \autocite{fasttext}. GloVe builds on Word2Vec by using global word statistics of the corpus during training, thus obtaining state-of-the-art performance on the word analogy dataset and producing vector spaces with meaningful sub-structure. The fastText method extends Word2Vec by using n-grams \keyword{n-gram} of characters instead of words. This approach helps to capture the significance of shorter words, suffixes, prefixes, and rare words.






\section{Transformers}
\label{sec:transformer}
\noindent
Word embedding models are great tools for obtaining token-level semantic vectors, but they offer little contextual information. It can be argued that an abstract representation of a phrase can be obtained by applying an operation (such as addition) to all the extracted vectors \autocite{w2v2}, but this is not a direct \textit{inductive bias}\keyword{inductive bias} of the model, but a property in the resulting vector space. Also, by doing something like summing all of the vectors in order to obtain a sequence representation, we lose information such as the order of the tokens. For tasks where sequentiality is important, as is the case in machine translation, this kind of information loss is not acceptable. A more expressive \textit{contextual embedding}\keyword{contextual embedding} is desirable. This contextual embedding should be able to represent an entire sequence of words or the contextual meaning of a word in a sequence, with respect to the other words and their order.

The modelling of sequential data was usually done using 1-dimensional Convolutional Neural Networks or \textit{Recurrent Neural Networks}\keyword{Recurrent Neural Network (RNN)} and the gated variants of the latter, such as the \textit{Long Short-Term Memory} \autocite{Hochreiter1997} and \textit{Gated Recurrent Units} \autocite{cho-etal-2014-learning} models\keyword{Long Short-Term Memory (LSTM), Gated Recurrent Unit (GRU)}. The problem with CNNs is that they have a limited context window, and some locality information is lost by applying operations such as pooling. RNNs, on the other hand, are able to encode a longer succession of tokens, but are notoriously hard to parallelize and can become very memory-expensive when training on long examples. Various tricks \autocite{rnn_tricks, ol-rnn} can help alleviate these issues, but the fundamental architectural constraints remain. 

A model that is able to circumvent these limitations is the Transformer \autocite{Vaswani2017}. Transformer networks are relinquishing recurrence in favour of the \textit{self-attention mechanism}\keyword{self-attention} to draw global dependencies between the input and the output. This approach enables much more parallelization, leading to much larger scale language models.

\begin{figure}[!t]
    \centering
    \includegraphics[width=.9\textwidth]{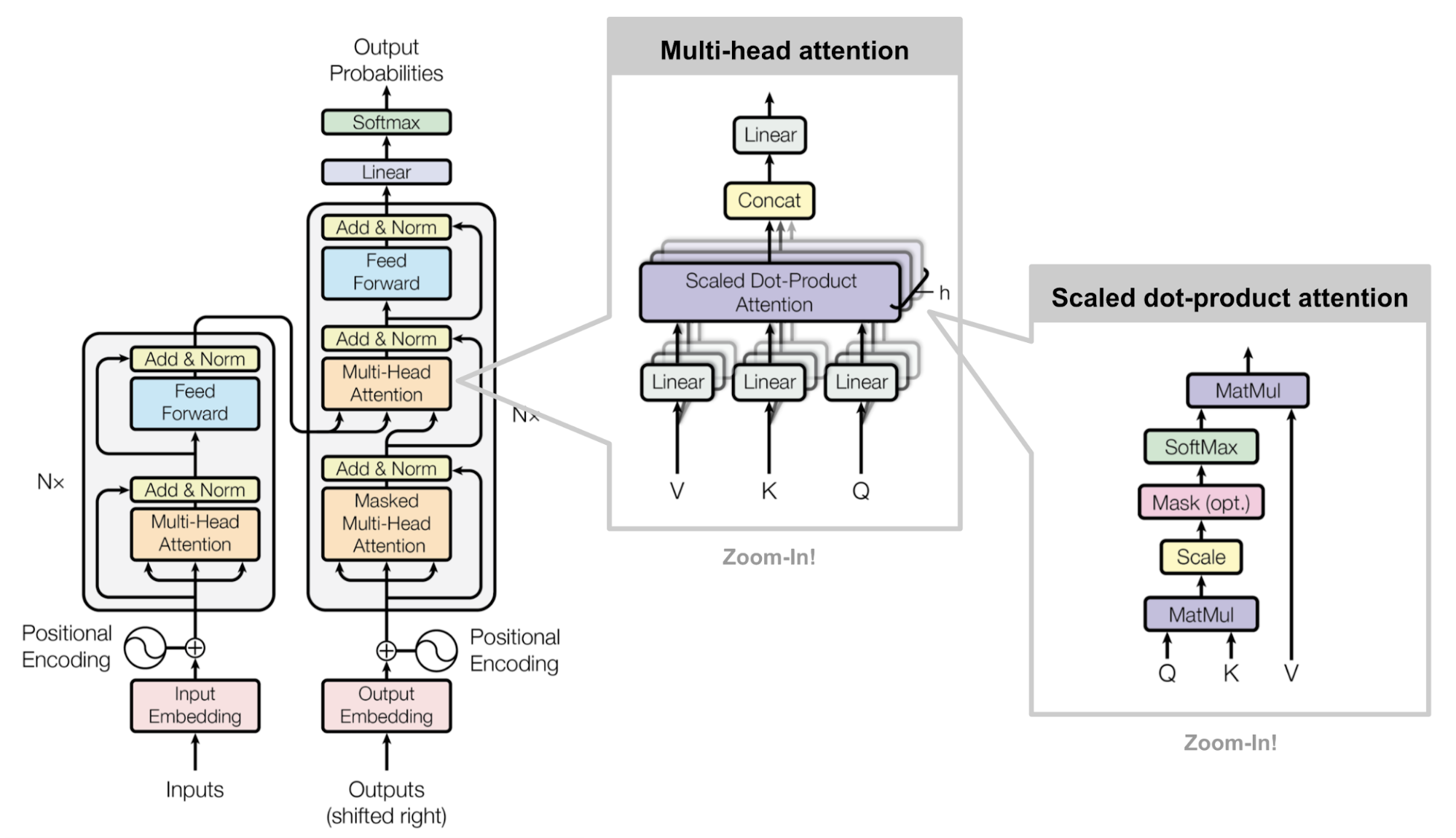}
    \caption{The original Transformer encoder-decoder architecture used for machine translation. The encoder layer has two sub-layers: a multi-head self-attention layer, and a fully-connected layer. One decoder layer contains two multi-head self-attention layers and a fully-connected layer. (figure from: \cite{weng_attn_attn, Vaswani2017})}
    \label{fig:transformer}
\end{figure}

The Transformer has an encoder-decoder structure, therefore it is able to map a sequence $x=(x_1, \dots, x_n)$ of symbols to a continuous representation $z=(z_1, \dots, z_n)$ which can be used in an \textit{auto-regressive}\keyword{auto-regressive} way to compute an output sequence of symbols $y=(y_1,\dots y_n)$. This encoder-decoder structure is constructed using just stacked self-attention and fully-connected layers, for both the encoder and the decoder, as can be seen in Fig. \ref{fig:transformer}.

The key component in the Transformer architecture is the multi-head self-attention mechanism. Self-attention is used to calculate a weighted average of the sequence representations, which are encoded into a \textit{key-value-query}\keyword{key, value, query (K, V, Q)} pair. The values of the (K, V, Q) pair are obtained by linearly projecting the word embeddings using a dot product between them and some learned weights (linear projection). A dot product between the query and the keys is then computed. The final weighted average is obtained by performing another dot product with the values matrix:

\begin{equation}
    Attention(Q, K, V) = softmax(\frac{QK^T}{\sqrt{d_k}})V
\end{equation}

\noindent
where $d_k$ is the dimension of the queries and keys. The scaling factor $\frac{1}{sqrt(d_k)}$ is used to prevent having large magnitude dot products, which could ``\textit{push the softmax functions into regions with extremely small gradients}''\sidenote{\textnormal{From \cite{Vaswani2017}: ``\textit{To illustrate why the dot products get large, assume that the components of $q$ and $k$ are independent random variables with mean $0$ and variance $1$. Then their dot product, $d\times k = \sum_{i=1}^{d_k}q_ik_i$, has mean $0$ and variance $d_k$}''}}.

This operation is done in parallel using multiple \textit{heads}, each head containing parameters for different linear projection matrices, thus allowing the model to process the information from different linear subspaces, resulting in more expressiveness. The final output is the following:

\begin{equation}
    MultiHead(X) = Concat(head_1,\dots,head_n)W^O
\end{equation}

\noindent where $head_i = Attention(XW_i^Q, XW_i^K,XW_i^V)$ and $X\in R^{batch \times tokens \times dim}$ is the input, $W_i^Q, W_i^K, W_i^V \in R^{d_{model}\times d_v}$ are the learnable linear projection matrices, and $W^O\in R^{hd_v\times d_{model}}$ is another learnable linear projection matrix. 

The output from the self-attention block can be then forwarded to a fully-connected network that's using a \textit{ReLU}\keyword{$ReLU(x)=max(0,x)$} activation function:

\begin{equation}
    FFN(x) = max(0,xW_1+b_1)W_2+b_2.
\end{equation}

The original Transformer model obtained state-of-the-art performances on multiple machine translation benchmarks while requiring fewer FLOPs to train than its counterparts. Models based on the Transformer architecture became very popular in natural language processing, especially when trained using self-supervised language modelling tasks, with the intention of being later fine-tuned for downstream tasks. Recently, Transformers have also been successfully used in computer vision \autocite{pic1616, imagegpt, dino}, and many various augmentations to the architecture have been proposed \autocite{reformer, longformer, nystromer}.

The reader should be aware that recent work in computer vision \autocite{mlp-mixer} suggests that self-attention is not needed and that one can design entirely fully-connected networks with similar performance to that of Transformers. However, this does not imply that self-attention and the Transformer are redundant, but merely that it's useful to also explore other inductive biases \autocite{no_free_lunch}.


\subsection{BERT}
\label{sub:bert}
\noindent
As mentioned in the previous subsection, a prevalent technique is to train large language models based on the Transformer and then use the language model for a specific (usually supervised) downstream task, by either extracting the representations or by fine-tuning the model on the new task.

One of the more successful models for accomplishing this is \textit{BERT}\keyword{Bidirectional Encoder Representations from Transformers (BERT)}. The authors argue that a key difference between BERT and previous techniques \autocite{gpt,elmo} is that the former ``\textit{restrict the power of the pre-trained representations}'' mainly because they are unidirectional language models, therefore limiting the attention to look only at the previous tokens during the pre-training phase, which could lead to sub-optimal results for sentence-level tasks. 

\begin{table}[!b]
    \centering
    \includegraphics[width=1\textwidth]{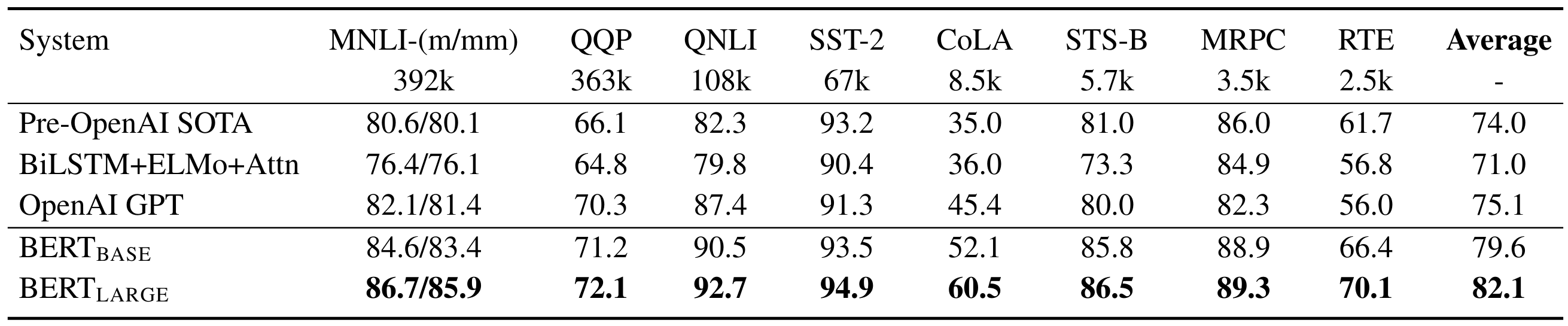}
    \caption{BERT scores on the GLUE test set. (table from: \cite{bert})}
    \label{fig:bert-glue}
\end{table}

BERT proposes two different tasks for pre-training: the masked language modelling task, which is inspired by the Cloze task \autocite{cloze}, and the \textit{next sentence prediction} \keyword{next sentence prediction (NSP)} task. The MLM task consists of randomly masking parts of the input and predicting the masked input with respect to the unmasked tokens. This allows for the self-attention mechanism to attend to the entire context during training, unlike the classical unidirectional language modeling task. The NSP task is a binary classification task for predicting whether the second sentence in a pair is a continuation of the first one. NSP is justified by the fact that it could improve the performance on various downstream tasks, such as question answering. However, further empirical research shows that the NSP task does not significantly improve downstream performance \autocite{roberta}. An illustration of the training and fine-tuning procedures for BERT can be consulted at Fig. \ref{fig:bert}.

The architecture of BERT consists of a multi-layer bidirectional Transformer encoder and the original model comes in two configurations: $BERT_{BASE}$ and $BERT_{LARGE}$. The first contains $12$ layers with a hidden size of $768$ and $12$ attention heads, with a total of $110M$ parameters, while the former contains $24$ layers with a hidden size of $1024$ and $16$ attention heads, totaling in $340M$ parameters. BERT is able to improve the state-of-the-art on $11$ NLP tasks, including GLUE (Table \ref{fig:bert-glue}) \autocite{glue}, SQuAD \autocite{squad}, and SWAG \autocite{SWAG}.

Many improvements and adaptations have been proposed over the original BERT. Some address other languages \autocite{camembert, bertje, robert}, domain-specific solutions \autocite{scibert, biobert}, or more efficient pre-training \autocite{albert, Clark2020}. We'll next present one such BERT-based model which leverages the idea of pre-training using two separate Transformers: a discriminator and a generator.

\subsection{ELECTRA}
\label{sub:electra}
\begin{figure}[!b]
    \centering
    \includegraphics[width=.8\textwidth]{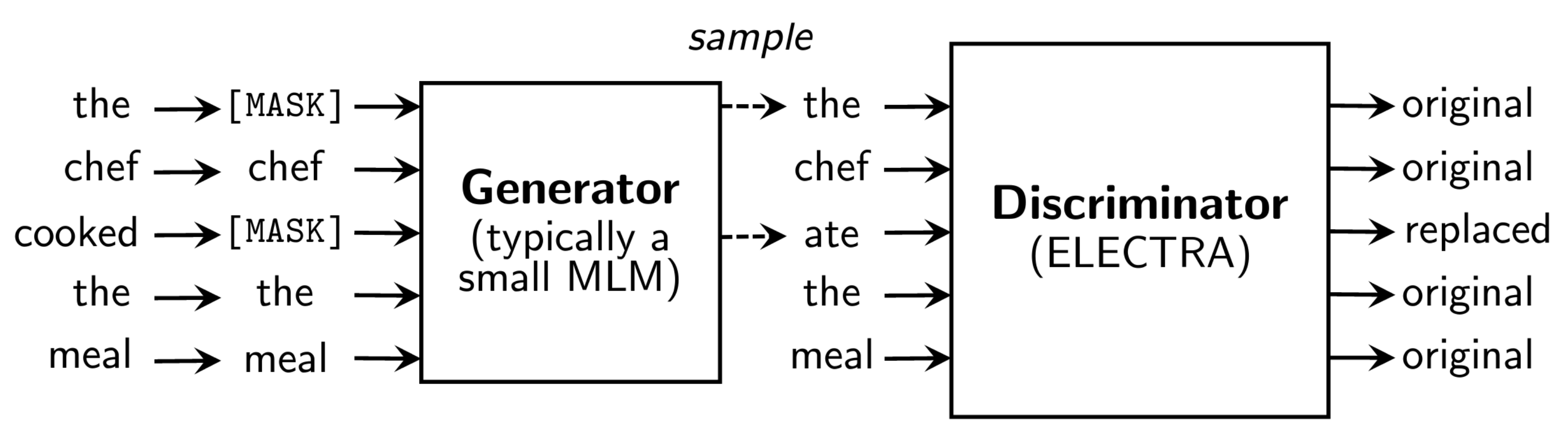}
    \caption{ELECTRA training scheme. (figure from: \cite{Clark2020})}
    \label{fig:ELECTRA}
\end{figure}

\noindent
\textit{ELECTRA}\keyword{Efficiently Learning an Encoder that Classifies Token Replacements Accurately (ELECTRA)} \autocite{Clark2020} is a method of pre-training by using a BERT-like generator and discriminator. The generator is trained with the MLM objective, as in BERT, and generates plausible alternatives for the masked tokens. The generated sequence is then fed to the discriminator which classifies every token as being \textit{original} or \textit{replaced} - this task is called replaced token detection (see Fig. \ref{fig:ELECTRA}, Fig. \ref{fig:date_train}). The network is trained using the following loss:

\begin{equation}
    \mathcal{L}_{ELECTRA} = \mathcal{L}_{MLM} + \lambda \mathcal{L}_{RTD}
\end{equation}

\noindent
where $\mathcal{L}_{MLM}$ is the masked language modeling loss:

\begin{equation}
    \mathcal{L}_{MLM} = \mathbb{E}[\sum_{i\in m}-log \mathbb{P}_G(x_i|\hat{x}(m); \theta_G)],
\end{equation}

\noindent
$\mathcal{L}_{RTD}$ is the replaced token detection loss:

\begin{equation}
    \mathcal{L}_{RTD} = \mathbb{E}[\sum_{i=1}^{T}-log\mathbb{P}_D(m_i|\Tilde{x}(m); \theta_D)],
\end{equation}

\noindent
$\theta_G$, $\theta_D$ are the generators' and discriminators' parameters, the set $m$ is the set of indices where the masking is applied, $\hat{x}(m)$ is the sequence without the replaced tokens at positions $m$, $\Tilde{x}(m)$ is the sequence with the tokens replaced at positions $m$, $\lambda$ is a loss weighting hyperparameter, $\mathbb{P}_G$ is the vocabulary probability distribution for each token, and $\mathbb{P}_D$ is the probability that a token was replaced or not. 

The generator is smaller than the discriminator, containing about $\frac{1}{4}$ of the discriminator parameters. The final model doesn't have shared parameters, except for the embedding layers. The reader should be aware of the fact that, even if the objective is similar to the one of a generative adversarial network, there are some key differences. First, the gradients from the discriminator are not propagating to the generator, mainly because there's a non-differentiable sampling phase when generating the corrupted input. Second, both the generator and the discriminator are trained using maximum likelihood, the method is thus distinct from the usual ``mini-max game'' played by the GAN discriminator and generator.

During training, $15\%$ of the tokens are masked. After training, the generator is discarded and only the discriminator is used for downstream tasks. Overall, this method greatly speeds up the training time. Furthermore, the fine-tuned discriminator is obtaining similar or better performance on all of the tasks (see Table \ref{fig:ELECTRA-glue}). 

\begin{table}[!t]
    \centering
    \includegraphics[width=1\textwidth]{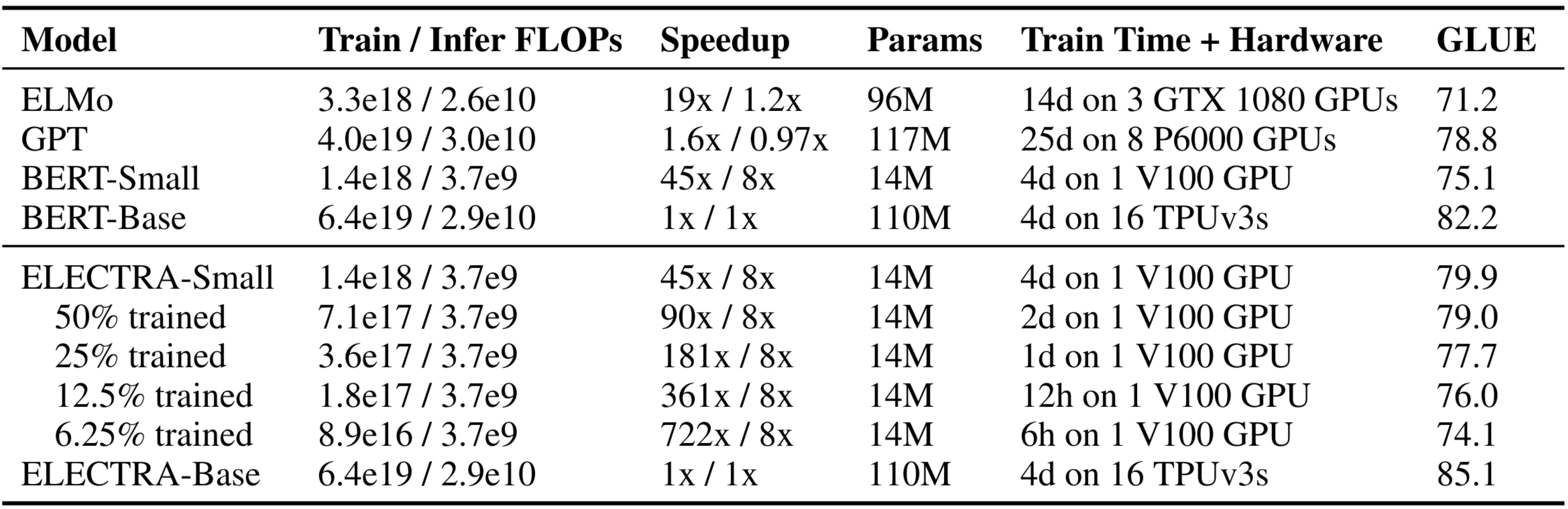}
    \caption{ELECTRA results on the GLUE dataset. (table from: \cite{Clark2020})}
    \label{fig:ELECTRA-glue}
\end{table}

A follow-up paper from the authors of ELECTRA explores the theoretical soundness of the training procedure \autocite{electric}. As a result, an \textit{energy-based} Cloze model\keyword{Energy-Based Model (EBM)} called Electric is developed, and it is argued that ELECTRA is a special case of Electric. This work is outside the scope of this thesis, but the author highly recommends that the interested reader should study both ELECTRA and Electric.



\chapter{Anomaly Detection}
\label{chp:anomalydetection}
\noindent
\paragraph{Many techniques} were proposed to address the outlier detection problem. If labels are available, we can treat the anomaly detection problem as an imbalanced supervised classification task and use common classification algorithms. However, data imbalance must be addressed. This is usually done by weighting the data or by adaptive resampling methods, such as SMOTE \autocite{smote}. The scenarios that interest us are the semi-supervised one, where we have access to clean data, and the unsupervised one, where we assume that our training set contains a percentage of outliers.

Some of the classic methods are based on \textit{density estimation} \keyword{density estimation, one-class learning, bayesian network, ensemble method, dictionary learning} \autocite{lof, elliptic_envelope, iso_forest}, one-class learning \autocite{ocsvm, Manevitz2001OneClassSF, svdd}, Bayesian networks \autocite{bayes3, bayes4}, ensemble methods \autocite{outlier-ensamble}, or approaches based on dictionary learning \autocite{irofti1, irofti2}.

Methods based on deep neural networks have become popular for AD due to their superior performance and the availability of large amounts of data. These methods usually involve learning the inlier class features using autoencoders \autocite{Hawkins2002, Sakurada2014, Chen2017} or generative adversarial networks \autocite{gan_anomaly, gan-ruff} and produce normality scores using the reconstruction error.

Recently, deep AD methods based on self-supervision have emerged in the field of computer vision, obtaining better quantitative and qualitative results. These approaches usually rely on training deep neural networks to distinguish between various transformations applied to the input images \autocite{Golan2018,e3outlier} and an anomaly score is computed by aggregating model predictions across multiple transformed input images. Such self-supervised approaches have also recently become common when dealing with video data \autocite{barbalau}. In the text domain, deep methods are more scarce and usually rely on pretrained word embeddings \autocite{acl2019, cvdd-2}.

In this chapter, we'll first describe two powerful classical baselines: the Isolation Forest \autocite{iso_forest} and the One-Class SVM \autocite{ocsvm}, and then we'll present two recent deep approaches - the \emph{E\textsuperscript{3} Outlier} \keyword{E\textsuperscript{3} Outlier}\autocite{e3outlier} framework, which enables the training of end-to-end self-supervised anomaly detectors by predicting transformations applied to the original images, and the \textit{Context Vector Data Description}\keyword{Context Vector Data Description (CVDD)} method, which is using pretrained word embeddings and self-attention to produce ``\textit{context vectors}'', which can be leveraged to produce anomaly scores for text sequences based on the \textit{cosine similarity}\keyword{cosine similarity} between the embeddings and the context vectors.

\section{Classical Models}
\label{sub:classic-od}
\subsection{One-Class Support Vector Machine}
\label{sub:ocsvm}
\noindent
Support Vector Machines \autocite{svm} are a class of models used to do binary classification. They achieve this by projecting the samples in a high-dimensional (or infinite dimensional) space, where the data is \textit{linearly separable}\keyword{linearly separable}, and choosing the decision boundary which has the \textit{maximum margin}\keyword{decision boundary, margin}. The concept of margin is defined as being the smallest distance between each sample and the decision boundary. Intuitively, this can be thought of as finding the hyperplane that separates the data into two classes and has the largest distance to all the nearest samples in the training data.

The One-Class SVM is a one-class classification technique based on the SVM. The strategy is to project the data into another space and then separate the data from the origin which has the maximum margin. By doing so, we can attribute the outlier class to the samples which are closer to the origin of the new space. Formally, let $\mathcal{X}$ be the initial feature space, and $\Phi:\mathcal{X}\to \mathcal{F}$ be a transformation that maps points from $\mathcal{X}$ to a space $\mathcal{F}$ which is equipped with a dot product that can be evaluated with a kernel function $k:\mathcal{X}^2\to\mathbb{R}$ which has the form:

\begin{equation}
    k(x,y) = (\Phi(x)\cdot \Phi(y)).
\end{equation}

To separate the data from the origin, we need to minimize the following expression:

\begin{equation}
    \mathcal{L} = \frac{1}{2}||W||^2 + \frac{1}{\nu N}\sum_{i=1}^{N} max\{b - W\cdot \Phi(x_i), 0\} - b,
\end{equation}

\noindent
where $W$ is the coefficient vector, $N$ is the number of samples in the training set, $\frac{1}{2}||W||^2$ is a regularization term, $b$ is the bias, $max\{b-W\cdot \Phi(X), 0\}$ is the so-called \textit{slack penalty}\keyword{slack penalty}, and $\nu$ is the term that regulates the trade-off between \textit{false positives} and \textit{false negatives}\keyword{false positive, false negative}. The term $\nu$ can be also be thought as being a ``\textit{prior probability that a data point in the training set is an outlier}'' \autocite{Aggarwal2015}.

A direct solution for the optimization problem is difficult to obtain due to the usage of $\Phi(\cdot)$, the projection function. Moreover, $\Phi(\cdot)$ is often not used directly when training SVMs, the transformation being implicitly done via the kernel function $k(\cdot, \cdot)$. One solution to this is by solving a \textit{dual representation}\keyword{dual problem, kernel trick} of the maximum margin problem, and using the \textit{kernel trick} within the dual formulation.

The dual problem is a constrained optimization problem and is outside the scope of this thesis. The reader should refer to \textit{Chapter 7.1} of Christoper Bishop's excellent book \autocite{bishop} for a more elaborate discussion regarding the optimization of SVMs. For the particular case of optimizing the OC-SVM, one can also refer to Chapter 3.4 of Charu C. Aggarwal's book \autocite{Aggarwal2015} on outlier detection.

\newpage

After training, an anomaly score can be obtained by computing:

\begin{equation}
    Score(x) = \sum_{i=1}^N \alpha_i\cdot k(x, x_i)-b,
\end{equation}

\noindent
where $\alpha_i$ is a Lagrangian parameter corresponding to the slack variables from the dual form.

\begin{figure}[!t]
    \centering
    \includegraphics[width=0.7\textwidth]{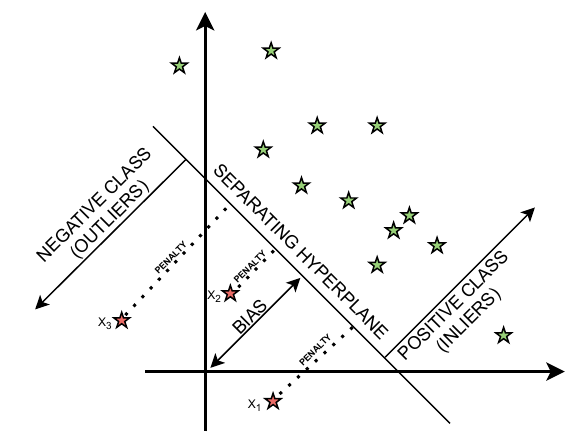}
    \caption{A solution to the OC-SVM optimization problem. Points on the left side of the hyperplane are classified as outliers, while points from on the right side are inliers. Outliers are penalized with $penalty=b-W\cdot\Phi(X_i)$. Inlier points are not penalized. (figure adapted from: \cite{Aggarwal2015})}
    \label{fig:ocsvm}
\end{figure}

\begin{figure}[!b]
    \centering
    \includegraphics[width=0.5\textwidth]{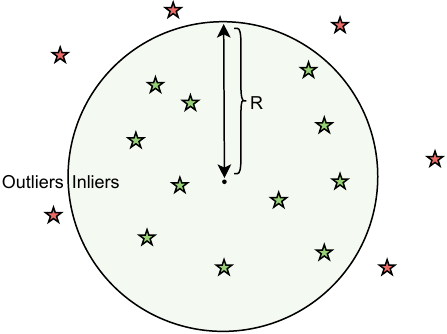}
    \caption{A SVDD solution. Inliers are inside the hypersphere of radius R, while outliers are outside of it.}
    \label{fig:svdd}
\end{figure}

A model that is closely related to the OC-SVM is the \textit{Support Vector Data Description} \autocite{svdd}\keyword{Support Vector Data Description (SVDD)}. The SVDD is an SVM that finds a hypersphere of radius $R$ in which the inliers are enclosed (as shown in Fig. \ref{fig:svdd}). This differs from the standard OC-SVM approach where a linear separator from the origin is found. The SVDD can be seen as a special case of the OC-SVM when using the \textit{Radial Basis Function}\keyword{Radial Basis Function (RBF)\\$k(x,y)=e^{ -\frac{||x-y||^2}{2\sigma^2} }$} kernel, which embeds the points on an unit sphere. Nonetheless, the two methods usually produce different solutions, and empirical evidence \autocite{svdd-better} suggests that the SVDD has a slight advantage over the OC-SVM using the RBF kernel.

\subsection{Isolation Forest}
\label{sub:isoforest}
\noindent
Instead of learning how the data distribution of inliers looks like, the \textit{Isolation Forest} \autocite{iso_forest}\keyword{Isolation Forest (iForest)} is identifying anomalous entries by isolating them into nodes and calculating an anomaly score based on the distance of the nodes from the root.

The Isolation Forest is built using an ensemble of \textit{Isolation Trees}\keyword{Isolation Tree (iTree)}, which are an unsupervised formulation of \textit{Decision Trees}\keyword{Decision Tree (DT)}. The iForest assumes that anomalies are a minority in the dataset which have attributes that greatly differ from the inlier class. iTrees are exploiting these properties of anomalies by recursively partitioning randomly selected attributes of the data with random cuts that are parallel to the axis. This has the effect of isolating outliers in shallow tree branches due to the points being located in sparse neighborhoods and thus making them more susceptible to isolation. The final decision is made by averaging the distances from the leaves to the root of the tree in different trees of the iForest. 

\begin{figure}[!b]
    \centering
    \includegraphics[width=0.8\textwidth]{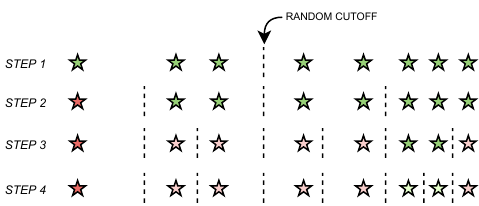}
    \caption{Isolation Tree partitioning process. The first data point is isolated at step 2 (higher anomaly score). All the points are isolated at the end of the recursive partitioning.}
    \label{fig:isoforest}
\end{figure}

The recursive partitioning done by an iTree is performed in the following way: given a sample of $n$ $d$-dimensional data points $X=\{x_1, \dots, x_n\}$, we consider $X$ as being the root node. We now divide $X$ by randomly picking an attribute $i$ and a random cutoff point $c\in [i_{min}, i_{max}]$, where $i_min$ and $i_max$ are the minimum and maximum values for attribute $i$. This gives us two subsets $X_{\leq} = \{x\in X | x^{(i)}\leq c \}$, and $X_> = \{ x\in X | x^{(i)} > c \}$. The two subsets represent the leaves of the initial node. We repeat this procedure recursively for every new node until either: (i) the tree reaches a height limit that was set priorly, (ii) the new subsets have exactly one element, or (iii) the new subsets contain the same value.  A toy example of this procedure is shown in Figure \ref{fig:isoforest}.

The average path length of an unsuccessful search in a Binary Search Tree\sidenote{\textnormal{the Isolation Tree is a Binary Search Tree}} is $c(n)=2H(n-1)-(2(n-1)/2)$, where $n$ is the number of external nodes, and  $H(n)\approx ln(n)+e$ is the harmonic number. The anomaly score for data point $x$ is defined as:

\begin{equation}
    Score(x, n) = 2^{ -\frac{\frac{1}{n_t} \sum_{i=1}^{n_t} h(x)}{c(n)}},
\end{equation}

\noindent
where $n_t$ is the number of iTrees in the Isolation Forest and $h(x)$ is the path length of a point measured by the number of edges that need to be traversed from the root to the point's corresponding node in the iTree.

The authors suggest that the following decisions can be made depending on the value of $Score$:

\begin{itemize}
    \item if $s(x, n)$ is close to $1$, then $x$ is an anomaly.
    \item if $s(x, n)$ is less than $0.5$, then $x$ can be regarded as an inlier.
    \item if $\forall x \in X$ we have $s(x,n)\approx 0.5$, then we are unable to tell if $X$ contains anomalies.
\end{itemize}

\section{Deep Models}
\label{sub:deep-od}
\subsection{E\textsuperscript{3} Outlier}
\label{sub:deep-od-cv}
\noindent
Some of the first artificial neural networks designed to tackle the anomaly detection problem were the ``\textit{replicator networks}'' \autocite{Hawkins2002}. The replicator network is an autoencoder trained to reconstruct the input data. The reconstruction error is then used to produce an anomaly score. Many recent deep AD methods rely on various autoencoder architectures \autocite{ae_od1, ae_od2}, but these kinds of networks suffer from the fact that autoencoders are generally not good at representation learning when compared to discriminative DNN architectures, such as ResNet \autocite{resnet}. This begs the question if discriminative networks could somehow be trained in an end-to-end fashion to do Outlier Detection.

The E\textsuperscript{3} (\textbf{E}ffective \textbf{E}nd-to-\textbf{E}nd) Outlier \autocite{e3outlier} is a neural network training and unsupervised anomaly detection framework for computer vision. Outlier detection is achieved by training the models via self-supervision\sidenote{\textnormal{In the original paper, the terms used are \textit{surrogate supervision} and \textit{surrogate supervision based discriminative network (SSD)}.}} with the task of discriminating between various transformations or combinations of transformations that are applied to the input data. An anomaly score is then calculated based on the model's predictions on multiple transformations or by using extracted representations in conjunction with other AD methods (such as the iForest).

\begin{figure}[!b]
    \centering
    \includegraphics[width=1\textwidth]{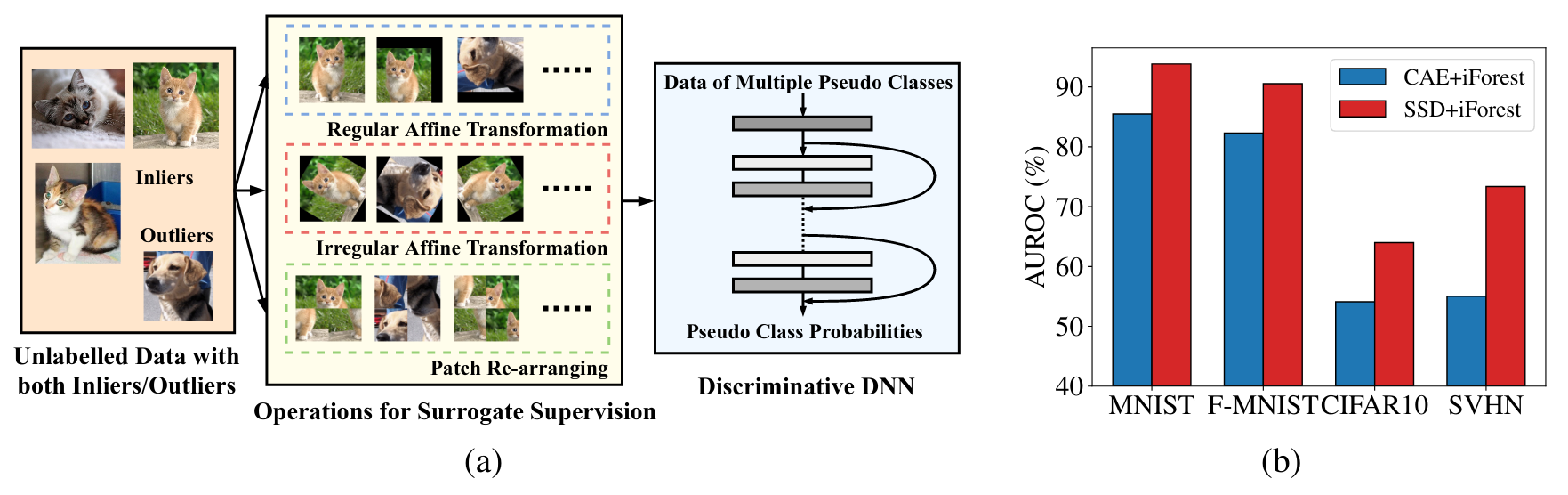}
    \caption{a) E\textsuperscript{3} Outlier self-supervision training schema. b) A performance comparison between representations extracted from the self-supervised discriminative network (SSD) and a convolutional autoencoder (CAE). \\(figure from: \cite{e3outlier})}
    \label{fig:e3_repr}
\end{figure}

The proposed self-supervision task can be defined in the following way: construct a set $\mathcal{O} = \{O(\cdot|y)\}_{y=1}^k$ of $K$ transformations that can be applied to the input images and label them with a label $y$. We can generate a new ``pseudo-labeled'' data point $x^{(y)}$ from an entry $x$ by applying the transformation $O(\cdot|y)\in \mathcal{O}$. Now we can train a SSD model which has a $K$-dimensional softmax layer in a self-supervised fashion by feeding it $x^{(y)}$, calculating the loss using \textit{cross entropy}\sidenote{$CE(\hat{y},y)=-\frac{1}{K}\sum_{i=1}^Ky_i\cdot log(\hat{y}_i)$}, backpropagating, and updating the neural network weights, as we would do with any supervised neural network. Please note that this technique requires no manually labeled data and that the ``pseudo-label'' is created at the moment in which we pick the transformation which we want to apply. We are not using any prior knowledge about the dataset itself.

The following sets of transformations are used:

\begin{enumerate}
    \item Rotations: the images are rotated clock-wisely by a certain degree.
    \item Flip: the images are flipped or not.
    \item Shifting: some pixels of the images are shifted along the $x$ or $y$ axis.
    \item Jigsaw: the images are partitioned into several equally-sized patches and then the patches are permuted.
\end{enumerate}

Based on these sets, we can define the ``regular affine transformations'' subset $\mathcal{O}_{RA}$, the ``irregular affine transformations'' subset $\mathcal{O}_{IA}$, and the ``patch re-arranging set'' $\mathcal{O}_{PR}$. These subsets can contain combinations of transformations, which have their own class. The final operation set used is $\mathcal{O}_{RA}\bigcup \mathcal{O}_{IA} \bigcup \mathcal{O}_{PR}$ and consists of $111$ transformations.

After the learning process, the learned representations can be used to train other anomaly detectors (as suggested in figure \ref{fig:e3_repr}b). Another method of obtaining an anomaly score is by directly utilizing the output from the softmax layer. Indeed, this approach produces better results when compared to the former. 

\begin{figure}[!b]
    \centering
    \includegraphics[width=.8\textwidth]{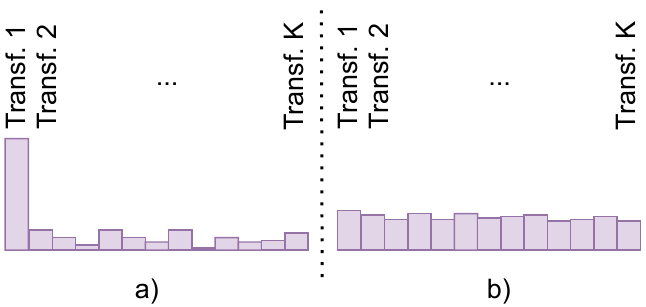}
    \caption{a) Low entropy ``spikey'' distribution. b) High entropy ``flat'' distribution. A high entropy gives a higher anomaly score.}
    \label{fig:entropy}
\end{figure}

To extract an anomaly score for an image $x$, we apply to every transformation $O(\cdot|y)$ in from the set $\mathcal{O}$, therefore obtaining a set of $K$ transformed images $\{x^{(1) \dots x^{K}}\}$ and their pseudo-labels. The transformed images are then fed through the network, producing a set of $K$ logits $\{ \mathbb{P}(x^{(y)}|\Theta) \}$, where $\Theta$ are the parameters of the network. Now we can apply one of the proposed scores:

\begin{enumerate}
    \item Pseudo Label (PL): an anomaly score is obtained by averaging the probabilities of the \textbf{correct class} from the obtained logits, i.e.:
    \begin{equation}
        PL(x) = \frac{1}{K}\sum_{y=1}^K\mathbb{P}^{(y)}(x^{(y)}|\Theta).
    \end{equation}
    
    \item Maximum Probability (MP): A issue with the PL score is that the classes obtained by applying the transformations could not be sufficiently separable. For example when we flip the number ``$8$'' we're still getting the number ``$8$''. This can make our models unable to detect the correct transformation and can hurt our anomaly score. One way to solve this issue is to take the \textbf{maximum probability from the logits} for every transformation:
    \begin{equation}
        MP(x) = \frac{1}{K} \sum_{y=1}^K \max_t \mathbb{P}^{(t)} (x^{(y)}|\Theta).
    \end{equation}
    
    \item Negative Entropy (NE): The previous two scores are using information from a single classes' probability, ignoring the other $(K-1)$ classes. One assumption that we can make is that, when given an anomaly, the neural network will be unable to decide what transformation was applied, thus producing an uninformative probability distribution that has high entropy, as seen in Fig. \ref{fig:entropy}. Under this assumption, we can use the entropy to produce an anomaly score in the following way:
    
    \begin{equation}
        NE(x) = -\frac{1}{K}\sum_{y=1}^K H(\mathbb{P}(x^{(y)} | \Theta)) 
    = \frac{1}{K}\sum_{y=1}^K\sum_{t=1}^K \mathbb{P}^{(t)}(x^{(y)}|\Theta)log(\mathbb{P}^{(t)}(x^{(y)}|\Theta)).
    \end{equation}
    
    In practice, the Negative Entropy score gives the best results.
\end{enumerate}

The authors are training multiple discriminative DNN models such as ResNet, Wide ResNet \autocite{wrn}, and DenseNet \autocite{densenet}, but the method can be applied to any neural network that can be trained on image data. The datasets used are MNIST \autocite{mnist}, Fashion MNIST \autocite{fmnist}, CIFAR10, SVHN \autocite{svhn}, and CIFAR100 \autocite{cifar}. The semi-supervised outlier detection setup is the following: given a dataset, consider that one class with one semantic concept (e.g. ``car'', ``human'') is the inlier class and use it as training data. For outliers, sample a percentage of the images from every other class. For the fully unsupervised scenario, simply add a percentage of outliers in the training dataset.

\begin{figure}[!t]
    \centering
    \includegraphics[width=1\textwidth]{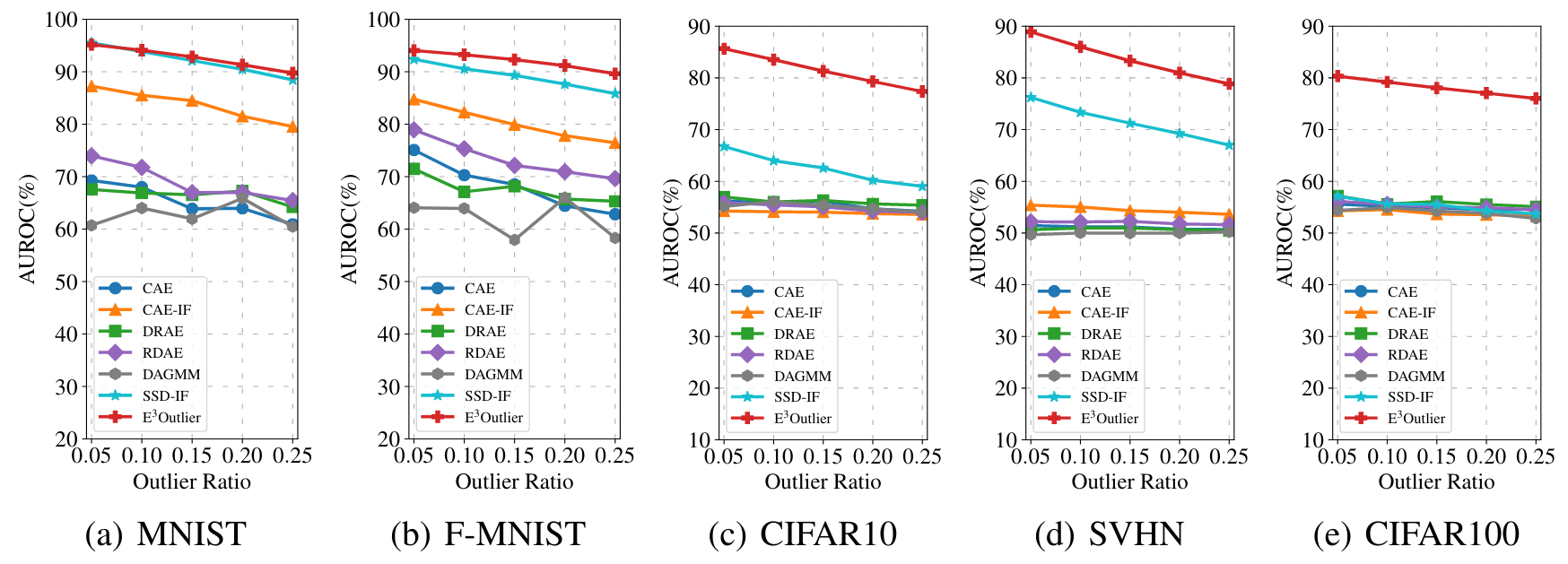}
    \caption{Unsupervised outlier detection comparison for various rates of contamination. (figure from: \cite{e3outlier})}
    \label{fig:uod-rates}
\end{figure}

The models trained in the E\textsuperscript{3} Outlier manner are greatly outperforming the competition, usually improving the \textit{AUROC} and \textit{AUPR}\keyword{Area Under the Receiver Operating Characteristic Curve (AUROC), Area Under the Precision-Recall Curve (AUPR)} scores by $5\%-30\%$ in both semi-supervised and unsupervised scenarios, as can be seen in Table \ref{fig:e3-selfsup} and Fig. \ref{fig:uod-rates}.

\begin{table}[!t]
    \centering
    \caption{AUROC/AUPR-inlier/AUPR-outlier (\%) scores for unsupervised outlier detection. Bold represents the best scores. $\rho$ is the training set contamination rate. CAE: Convolutional autoencoder. CAE-IF: CAE+iForest. DRAE: Discriminative reconstruction based autoencoder. RDAE: Robust deep autoencoder. DAGMM: Deep autoencoding gaussian mixture model (table from: \cite{e3outlier})}
    \includegraphics[width=1\textwidth]{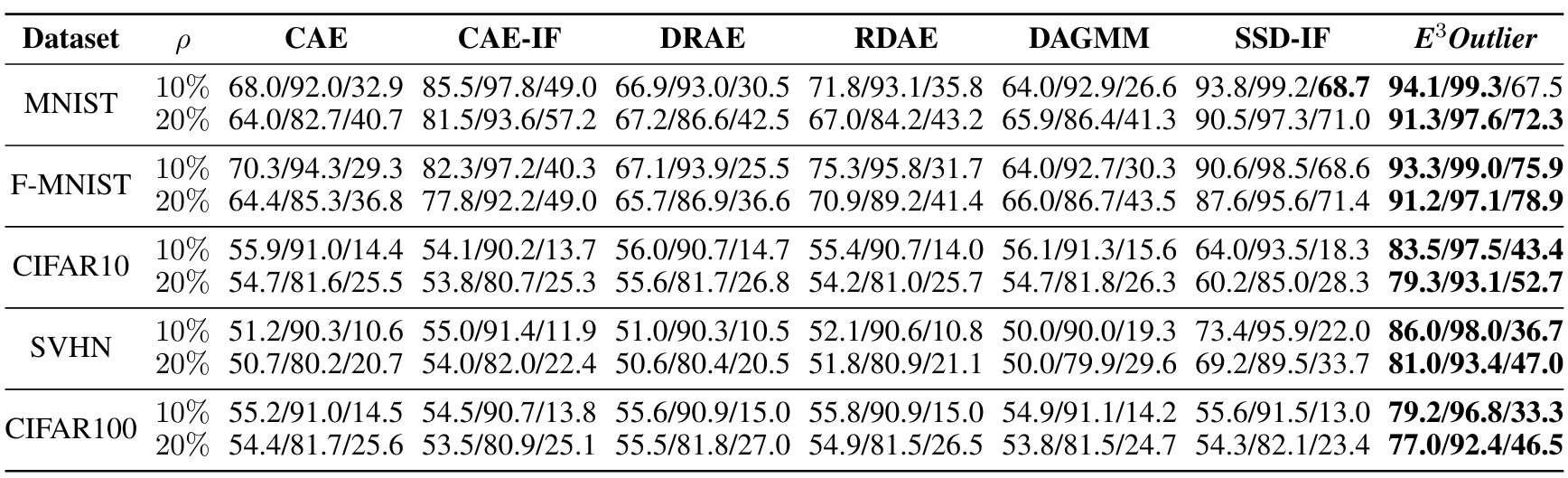}
    \label{fig:e3-selfsup}
\end{table}

\subsection{Context Vector Data Description}
\label{sub:deep-od-nlp}
\noindent
While many anomaly detection solutions using deep neural networks were specifically designed for computer vision, only a few have been proposed for natural language processing. One method for detecting anomalies in text is the Context Vector Data Description \autocite{acl2019}, which leverages pretrained word embeddings in order to construct ``\textit{context vectors}''\keyword{context vector} using the self-attention mechanism. These context vectors can capture semantic contexts and can be used to perform contextual anomaly detection with respect to the themes and concepts which are encoded in the learned context vectors.

The CVDD method is using a slightly different multi-head self-attention mechanism than the one presented in \autocite{Vaswani2017}. For a given sentence $S=(w_1, \dots, w_l)$, we can extract a matrix $H=(h_1, \cdot, h_l)\in \mathbb{R}^{p \times l}$ containing the corresponding word embeddings, extracted with some universal word embedding models (e.g. Word2Vec, FastText, GloVe), or by some language model (e.g. BERT, ELECTRA). Now, by using the multi-head self-attention mechanism, we can obtain a single fixed-size representation from our matrix $H$. First, we compute the attention matrix $A$ in the following way:

\begin{equation}
    A = softmax(tanh(H^TW_1)W_2),
\end{equation}

\noindent
where $W1\in\mathbb{R}^{p\times d_a}$ and $W_2\in \mathbb{R}^{d_a\times r}$ are learnable weight matrices, and $d_a$ is the model dimension. The \textit{tahn}\sidenote{$tanh(x)=\frac{e^x-e^{-x}}{e^x+e^{-x}}$} activation function is applied element-wise while the softmax activation is applied column-wise. By doing this we obtain an attention matrix $A=(a_1,\dots,a_r)$ containing $r$ attention heads. Now, a sentence embedding matrix $M=(m_1,\dots,m_r)\in\mathbb{R}^{p\times r}$ can be calculated using the self-attention matrix:

\begin{equation}
    M = HA.
\end{equation}

\begin{figure}[!b]
    \centering
    \includegraphics[width=1\textwidth]{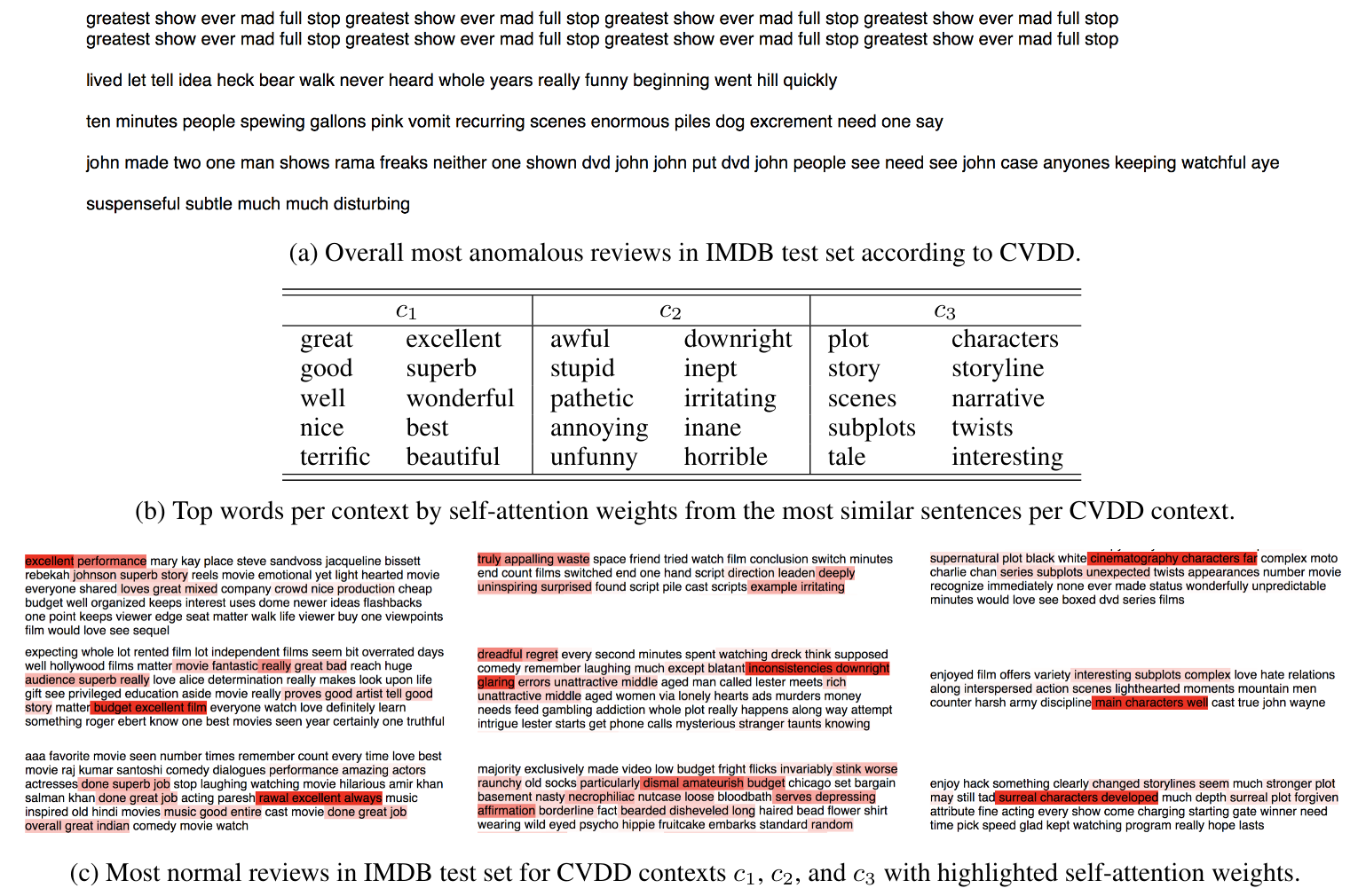}
    \caption{Qualitative results on the IMDB Movie Reviews dataset. a) The most anomalous entries in the test set. b) Words that are clustering around the context vectors - the first and second context vectors seem to model positive and negative sentiments, while the third one models cinematic language. c) Top most normal entries in the test set. Most abnormal words are in a more intense red. (figure from: \cite{acl2019})}
    \label{fig:cvdd_imdb}
\end{figure}

The context vectors can now be constructed as follows: let $r$ be the number of attention heads and $C=(c_1,\dots,c_r)\in \mathbb{R}^{p\times r}$ a matrix containing $r$ vectors. We say that $C$ is the context matrix and $c_1,\dots,c_r$ are the context vectors. The Context Vector Data Description loss is defined in the following way:

\begin{equation}
    \min_{C, W_1, W_2}\frac{1}{n} \sum_{i=1}^n\sum_{k=1}^r \sigma_k(H^{(i)})d(c_k, m_k^{(i)}) + ||C^TC-I||_F^2,
\end{equation}

\noindent
where $d(c_k, m_k^{(i)})$ is the cosine distance between the context vector $c_k$ and the data representation $m_k^{(i)}$, $I$ is the $r\times r$ identity matrix, $||\cdot||_F$ is the Frobenius norm, and $\sigma_1(H), \dots, \sigma_r(H)$ are the input-dependent weights which are passed through a \textit{parametrized softmax function}\sidenote{$\sigma_k(H)=\frac{e^{-\alpha d(c_k, m_k(H))}}{ \sum_{j=1}^r e^{ -\alpha d(c_j, m_j(H)) } }$}. Note that there are as many context vectors as attention heads, thus constraining the network to learn attention weights which are able to extract the most common themes and concepts from the given corpora. The $||C^TC-I||_F^2$ regularization term is also used to promote diverse context vectors, by encouraging orthogonality. The network is trained using gradient descent, with respect to the weights $W_1, W_2$, and the context matrix $C$.

After training, an anomaly detection score can be calculated for a sample $S$ with an embedding $H$ by averaging the contextual anomaly scores:

\begin{equation}
    Score(H) = \frac{1}{r}\sum_{k=1}^r d(c_k, m_k(H)).
\end{equation}

\begin{table}[!t]
    \centering
    \includegraphics[width=1\textwidth]{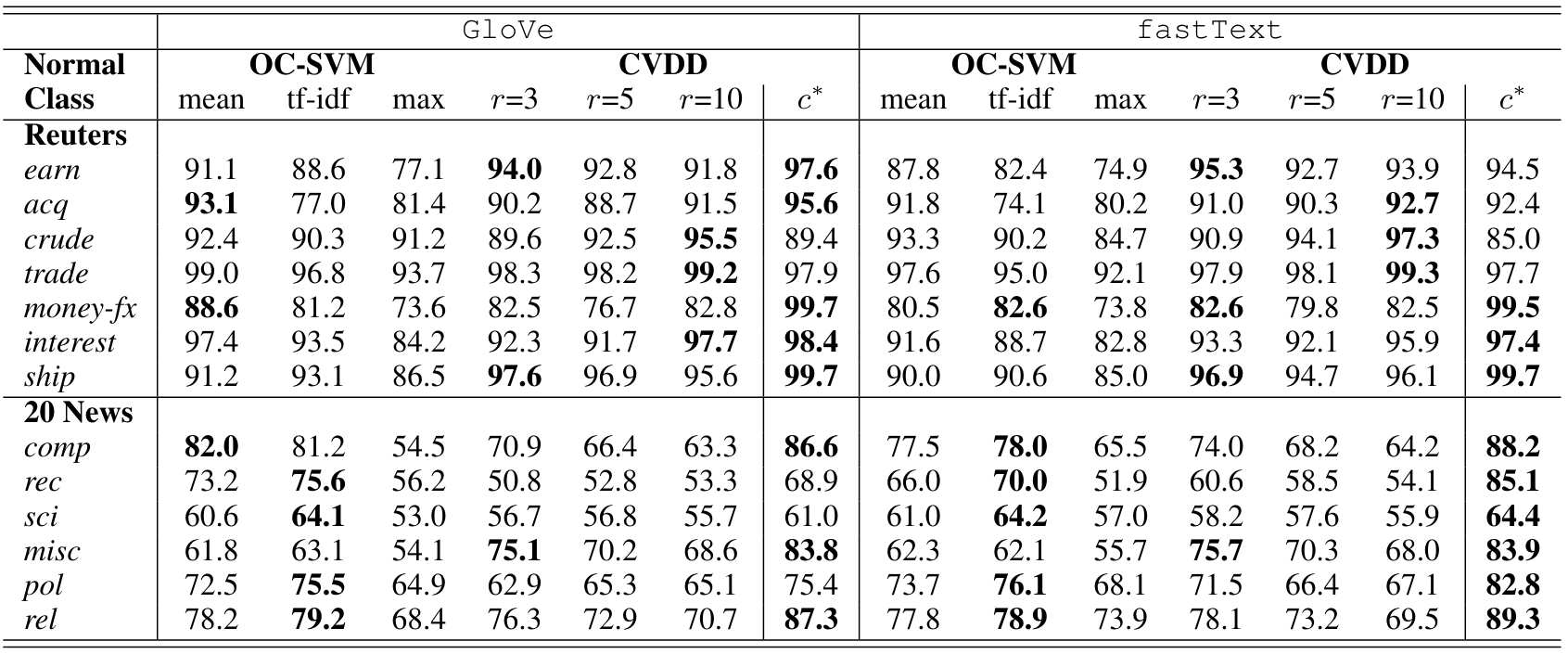}
    \caption{Semi-supervised outlier detection scores (AUROC \%) on the Reuters and 20 Newsgroups datasets. The $c*$ context vector is hand-picked and produces the best scores, thus making it dependent on the ground truth. (table from: \cite{acl2019})}
    \label{fig:cvdd1}
\end{table}

Quantitative and qualitative experiments are performed on 20Newsgroups \autocite{20ng}, Reuters-21578 \autocite{reuters} (See Table \ref{fig:cvdd1} and Table \ref{fig:cvdd2}), and IMDB Movie Reviews \autocite{imdb} datasets (See Fig. \ref{fig:cvdd_imdb}). The official train/test split is used. One class is considered as being inliers, while the rest are considered outliers.

The word embeddings used are GloVe and fastText. On every dataset, the text is stripped of punctuation, numbers, redundant whitespaces, and stopwords. The text is also converted to lowercase and the only words that are considered are the words with more than three characters.

\begin{table}[!b]
    \centering
    \includegraphics[width=1\textwidth]{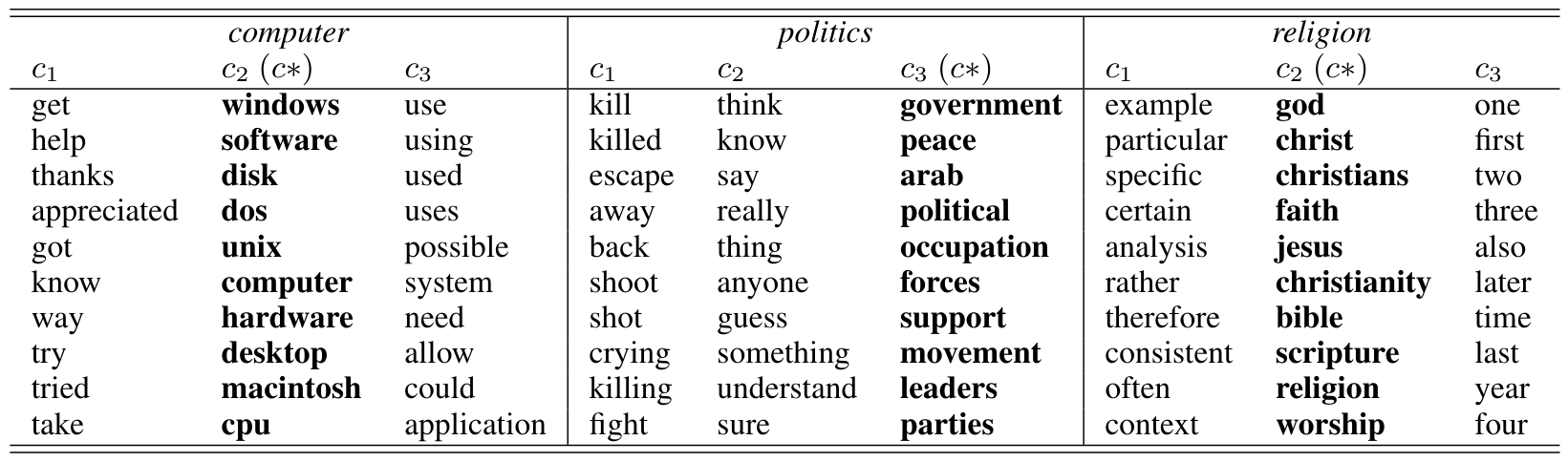}
    \caption{Top words per context on the 20 Newsgroups comp, pol, and rel subsets with 3 context vectors. (table from: \cite{acl2019})}
    \label{fig:cvdd2}
\end{table}

\chapter{DATE}
\label{chp:date}
\noindent
\paragraph{As announced}, in this chapter we shall introduce our solution to the problem of detecting anomalies in text using deep learning. Our method is called DATE (which is short for \textbf{D}etecting \textbf{A}nomalies in \textbf{T}ext using \textbf{E}LECTRA) \autocite{DATE}. Our solution is an end-to-end AD method relying on self-supervision in order to produce an anomaly score. We will first concern ourselves with the RMD and RTD self-supervision tasks which were also described in the Prologue chapter of the thesis.

\section{Pretext tasks}
\label{sub:pretext}
\noindent
The discriminative task called RMD has the purpose of creating training data by altering a given text using one out of $K$ given operations and it also asks to ascertain which one of the operations is the correct one given the transformed text. The modification of the text is obtained by following the next steps:
\begin{enumerate}
    \item \textit{masking} some of the input words using a predefined mask pattern;
    \item replacing the \textit{masked} words with alternative ones\sidenote{\textnormal{For example \textit{frog} with \textit{animal}}}.
\end{enumerate}

We shall now proceed to describe our method more formally. To this end, let $\bm{m} \in \{0, 1\}^T$ be a mask pattern corresponding to the text input $\bm{x} = [x_1, \dots, x_T].$ For training, we generate and fix $K$ mask patterns $\bm{m}^{(1)}, \dots, \bm{m}^{(K)}$ by randomly sampling a constant number of $1'$s out of the $T$ available ones. Unlike ELCTRA which masks random tokens spontaneously, our model first samples a mask model from the $K$ predefined ones. We then proceed to mask our input with $\bm{m}$ and obtain the vector $\bm{\hat{x}}(m) := [\hat{x}_1, \dots, \hat{x}_T],$ where $$\hat{x}_i = \begin{cases}
x_i, & \bm{m}_i = 0 \\
\texttt{[MASK]}, & \bm{m}_i = 1\end{cases}$$
or simply $\hat{x}_i = x_i (1 - \bm{m}_i) + \texttt{[MASK]} \bm{m}_i$. For instance, given an input $\bm{x} = [$they, were, ready, to, go$]$ and a mask pattern $\bm{m} = [0, 0, 1, 0, 1],$ the resulting masked input will be $\hat{\bm{x}}(\bm{m}) = [$they, were, $\texttt{[MASK]}$, to, \texttt{[MASK]}$].$

Each of the masked tokens could then be replaced by a word token. A naive way to do this is by sampling uniformly from the vocabulary. However, for more plausible alternatives, masked tokens can be sampled from a Masked Language Model generator such as BERT, which outputs a probability distribution $\mathbb{P}_G$ over the vocabulary for each corrupted token. We will denote the plausibly corrupted text by $\widetilde{\bm{x}}(\bm{m}) = [\widetilde{x}_1, \dots, \widetilde{x}_T],$ where
$$\widetilde{x_i}=
    \begin{cases}
      x_i, & \bm{m}_i=0 \\
      w_i\sim \mathbb{P}_{G}(x_i|\bm{\hat{x}}(\bm{m});\theta_G), & \bm{m}_i=1
\end{cases}$$
For instance, for the masked input from before, a plausibly corrupted input could be $\widetilde{\bm{x}}(\bm{m}) =$ [they, were, prepared, to, depart].

RTD is a binary sequence tagging task, where some tokens in the input are corrupted with plausible alternatives, which is similar to how RMD works. The key difference is that the discriminator must then predict for \textit{each token} if it's the \emph{original} token or a \emph{replaced} one. Distinctly from RTD, which is a \emph{token-level} discriminative task, RMD is a \emph{sequence-level} one. This means that the model distinguishes between a fixed number of predefined transformations applied to the input. As such, RMD can be seen as the text counterpart task for the self-supervised classification of geometric alterations applied to images, as described in Subsec. \ref{sub:deep-od-cv}. While RTD predictions could be used to sequentially predict an entire mask pattern, they can lead to masks that are not part of the predefined $K$ patterns\sidenote{\textnormal{Suppose that we have the input text ``\textit{the cat is purring}'' and the masking pattern set $\mathcal{M}=\{[0,1,0,1], [1,0,0,1]\}$. In this situation, we could use the RTD head to produce both available masking patterns, but it could also produce the pattern $[1,1,1,1]$, which does not exist in $\mathcal{M}$.}}. The RMD constraint overcomes this behaviour. We thus train DATE to solve both tasks simultaneously, which increases the AD performance compared to solving one task only, as shown in Subsec. \ref{sec: ex-ablation}. Furthermore, we observed empirically that this approach also improves training stability.

\subsection{DATE Architecture}
\label{sub:DATEarch}
\noindent
We propose simultaneously solving RMD and RTD by jointly training a generator, \textbf{G}, and a discriminator, \textbf{D}.

\begin{figure}[t!]
\centering
\includegraphics[width=0.7\columnwidth]{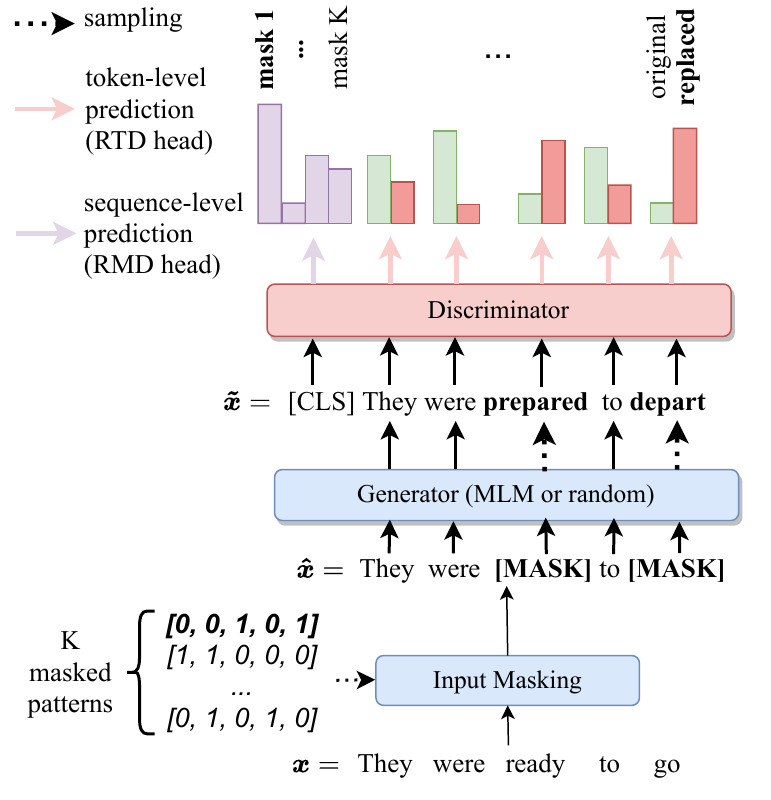}
\caption{DATE Training. Firstly, the input sequence is masked using a sampled masked pattern and a generator fills in new tokens in place of the masked ones. Secondly, the discriminator receives supervision signals from two tasks: RMD (which mask pattern was applied to the input sequence) and RTD (the per-token status: original or replaced). (figure from: \cite{DATE})}
\label{fig:date_train}
\end{figure}

\textbf{G} is a MLM used to \emph{replace} the masked tokens with plausible alternatives. We also consider a setup with a \emph{random generator}, in which we sample tokens uniformly from the vocabulary. Indeed, this approach seems to work better for our AD setup than a parameterized generator, as can be seen in the ablation study at Subsec. \ref{sec: ex-ablation}. We suspect that this is happening because the anomalies are a lot more different than the inliers, thus making the plausible text generated by a parameterized generator look much like the inlier class, ultimately confusing the discriminator.  

\textbf{D} is a deep neural network with two prediction heads used to distinguish between \emph{corrupted} and original tokens (RTD) and to predict which mask pattern was applied to the corrupted input (RMD). At test time, \textbf{G} is discarded, and \textbf{D}'s probabilities are used to compute an anomaly score.

Both \textbf{G} and \textbf{D} models are based on a BERT encoder, which consists of several stacked Transformer blocks. The BERT encoder transforms an input token sequence $\bx~=~[x_1, x_2, ..., x_T]$ into a sequence of contextualized word embeddings $h(\bx) = [h_1, h_2, ..., h_T]$.

When parameterized, \textbf{G} is a BERT encoder with a linear layer on top that outputs the probability distribution $P_G$ for each token. The generator is trained using the MLM loss:
\begin{equation}
    \mathcal{L}_{MLM}= \E\biggl[\sum_{\substack{i=1;\\s.t. m_i=1}}^{T} -\log P_G(x_i|\bm{\hat{x}}(\bm{m});\theta_G)\biggr]
\end{equation}

\textbf{D} is a BERT encoder with two prediction heads applied over the contextualized word representations:

\begin{enumerate}
    \item \textbf{RMD head.} This head outputs a vector of logits for all mask patterns $\bm{o}=[o_1,...,o_K]$. We use the contextualized hidden vector $h_{\texttt{[CLS]}}$ (corresponding to the $\texttt{[CLS]}$ special token at the beginning of the input) for computing the mask logits $\bm{o}$ and $P_M$, the probability of each mask pattern:
    \begin{equation}
        \mathbb{P}_{M}(\bm{m}=\bm{m}^{(k)}|\,\widetilde{\bm{x}}(\bm{m}^{(k)}); \theta_D)=\frac{e^{o_k}}{\sum_{i=1}^{K} e^{o_i}}
    \end{equation}
    \item \textbf{RTD head.} This head outputs scores for the two classes (\emph{original} and \emph{replaced}) for each token $x_1, x_2, ..., x_T$, by using the contextualized hidden vectors $h_1, h_2, ..., h_T$.
\end{enumerate}

We train the DATE network in a maximum-likelihood fashion using the $\mathcal{L}_{DATE}$ loss:
\begin{equation}
     \min_{\theta_D, \theta_G}  \sum_{\bm{x} \in \mathcal{X}} \mathcal{L}_{DATE}(\theta_D,\theta_G; \bm{x})
\end{equation}

 The loss contains both the token-level losses in ELECTRA, as well as the sequence-level mask detection loss $\mathcal{L}_{RMD}$:
\begin{align}
\begin{split}
    \mathcal{L}_{DATE}(\theta_D,\theta_G;\bm{x}) = \mu\mathcal{L}_{RMD}(\theta_D;\bm{x}) + \mathcal{L}_{MLM}(\theta_G;\bm{x}) + \lambda\mathcal{L}_{RTD}(\theta_D;\bm{x}),
\end{split}
\end{align}
where the discriminator losses are:
\begin{align}
    \mathcal{L}_{RMD}&=\E \biggl[-\log \mathbb{P}_{M}(\bm{m}|\,\widetilde{\bm{x}}(\bm{m});\theta_D)\biggr], \\
    \mathcal{L}_{RTD}&=\E\biggl[ \sum_{\substack{i=1;\\x_i\ne \texttt{[CLS]}}}^{T} -\log \mathbb{P}_D(m_i|\,\bm{\widetilde{\bm{x}}}(\bm{m});\theta_D)\biggr],
\end{align}
where $\mathbb{P}_D$ is the probability distribution that a token was replaced or not.


The ELECTRA loss enables \textbf{D} to learn good feature representations for language understanding. Our RMD loss puts the representation in a larger sequence-level context. After pre-training, \textbf{G} is discarded and \textbf{D} can be used as a general-purpose text encoder for downstream tasks. Output probabilities from \textbf{D} are further used to compute an anomaly score for new examples.

\subsection{Anomaly Detection score}

\begin{figure}[b!]
\centering
\includegraphics[width=0.7\columnwidth]{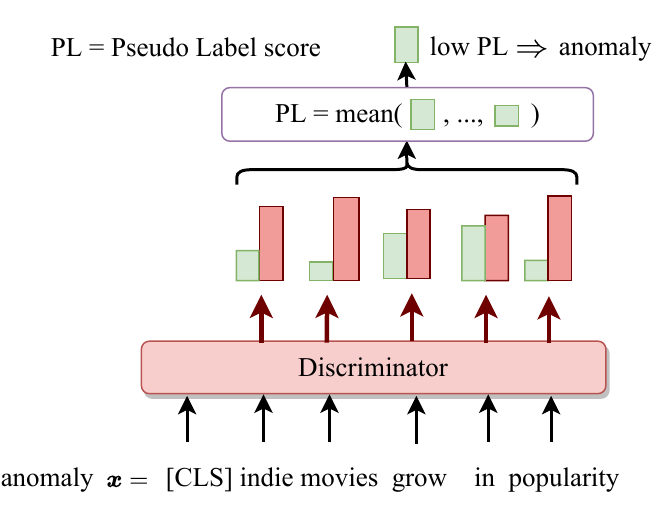}
\caption{DATE Testing. The input text sequence is fed to the discriminator, resulting in token-level probabilities for the normal class, which are further aggregated into an anomaly score, as detailed in Sec.\ref{sec:pl_score}. For deciding whether a sample is either normal or abnormal, we aggregate over all of its tokens. (figure from: \cite{DATE})}
\label{fig:date_test}
\end{figure}

\label{sec:pl_score}
We adapt the Pseudo Label (PL) based score from the \emph{$E^{3}Outlier$} framework in a novel and efficient way. In its general form, the PL score aggregates responses corresponding to multiple transformations of $\bx$. This approach requires $K$ input transformations over an input $\bx$ and $K$ forward passes through a discriminator. It then takes the probability of the ground truth transformation and averages it over all $K$ transformations. 

To compute PL for our RMD task, we take $\bx$ to be our input text and the K mask patterns as the possible transformations. We corrupt $\bx$ with mask $\bm{m}^{(i)}$ and feed the resulted text to the discriminator. We take the probability of the i\textsuperscript{th} mask from the RMD head. We repeat this process $K$ times and average over the probabilities of the correct mask pattern. This formulation requires $K$ feed-forward steps through the DATE network, which slows down inference. We propose a more computationally efficient approach next.

Instead of aggregating \emph{sequence-level} responses from multiple transformations over the input, we can aggregate \emph{token-level} responses from a single model over the input to compute an anomaly score. More specifically, we can discard the generator and feed the original input text to the discriminator directly. We then use the probability of each token being \emph{original} (not \emph{corrupted}) and then average over all the tokens in the sequence:
\begin{equation}
    \label{eq: PL_score}
    PL_{RTD}(x) = \frac{1}{T} \sum_{i=1}^T
    P_D(m_i=0|\bm{\widetilde{\bx}}(\bm{m}^{(0)});\theta_D),
\end{equation}
where $\bm{m}^{(0)}=[0, 0, ..., 0]$ effectively leaves the input unchanged.
As can be seen in Fig.~\ref{fig:date_test}, the RTD head will be less certain in predicting the \emph{original} class for \emph{outliers} (having a probability distribution unseen at training time), which will lead to lower PL scores for \emph{outliers} and higher PL scores for \emph{inliers}. We use PL at testing time, when the entire input is either normal or abnormal. Our method also speeds up inference, since we only do one feed-forward pass through the discriminator instead of $k$ passes. Moreover, having a per token anomaly score helps us better understand and visualize the behavior of our model, as shown in Fig.~\ref{fig: qual_ex}.

\section{Experimental analysis}
\noindent
In this section, we detail the empirical validation of our method by presenting: the semi-supervised and unsupervised experimental setup, a comprehensive ablation study on DATE, and the comparison with state-of-the-art on the semi-supervised and unsupervised AD tasks. DATE does not use any form of pre-training or knowledge transfer (from other datasets or tasks), learning all the embeddings from scratch. Using pre-training would introduce unwanted prior knowledge about the outliers, making our model considering them known (normal).

\subsection{Experimental setup}
\noindent
We describe next the Anomaly Detection setup, the datasets and the implementation details of our model. The code is publicly available~\sidenote{\url{https://github.com/bit-ml/date}}.

We use a semi-supervised setting in Sec.~\ref{sec: ex-ablation}-\ref{sec: ex-ssad} and an unsupervised one in Sec.~\ref{sec: ex-uad}. In the semi-supervised case, we successively treat one class as normal (\emph{inliers}) and all the other classes as abnormal (\emph{outliers}). In the unsupervised AD setting, we add a fraction of outliers to the inliers training set, thus contaminating it. We compute the Area Under the Receiver Operating Curve (AUROC) for comparing our method with the previous state-of-the-art. For a better understanding of our model's performance in an unbalanced dataset, we report the Area Under the Precision-Recall curve (AUPR) for inliers and outliers per split in the supplementary material~\ref{apx:prauc_experiments}.

We test our solution using two text classification datasets, after stripping headers and other metadata. For the first dataset, 20Newsgroups, we keep the exact setup, splits, and preprocessing (lowercase, removal of: punctuation, number, stop words, and short words) as in \autocite{acl2019}, ensuring a fair comparison with previous text anomaly detection methods. As for the second dataset, we use a significantly larger one, AG News, better suited for deep learning methods.

\begin{enumerate}
    \item \textbf{20Newsgroups}\sidenote{\url{http://qwone.com/~jason/20Newsgroups/}}: We only take the articles from six top-level classes: \emph{computer, recreation, science, miscellaneous, politics, religion}, like in \autocite{acl2019}. This dataset is relatively small, but a classic for NLP tasks (for each class, there are between 577-2856 samples for training and 382-1909 for validation).
    
    \item \textbf{AG News} \autocite{ag_news}: This topic classification corpus was gathered from multiple news sources, for over more than one year \sidenote{\url{http://groups.di.unipi.it/~gulli/AG_corpus_of_news_articles.html}}. It contains four topics, each class with 30000 samples for training and 1900 for validation.

\end{enumerate} 

For training the DATE network we follow the pipeline in Fig.~\ref{fig:date_train}. In addition to the parameterized generator, we also consider a \emph{random generator}, in which we replace the masked tokens with samples from a uniform distribution over the vocabulary.

The \emph{discriminator} is composed of four Transformer layers, with \emph{two prediction heads} on top (for RMD and RTD tasks). We provide more details about the model in the supplementary material~\ref{apx:model_implementation_details}. We train the networks with AdamW with amsgrad~\autocite{adamw-amsgrad}, $1e^{-5}$ learning rate, using sequences of maximum length $128$ for AG News, and $498$ for 20Newsgroups. We use $K=50$ predefined masks, covering $50\%$ of the input for AG News and $K=25$, covering $25\%$ for 20Newsgroups. The training converges on average after $5000$ update steps and the inference time is $0.005$ sec/sample in PyTorch~\autocite{pytorch}, on a single GTX Titan X.

\subsection{Ablation studies}
\label{sec: ex-ablation}
\noindent
To better understand the impact of different components in our model and making the best decisions towards a higher performance, we perform an extensive set of experiments (see Tab.~\ref{tab: ablation}). Note that we successively treat each AG News split as inlier and report the mean and standard deviations over the four splits. The results show that our model is robust to domain shifts.

\begin{table}[t!]
\begin{center}
	\begin{tabular}{r r l c}
		\toprule
		\multicolumn{1}{p{0.7cm}}{Abl.} &
        \multicolumn{1}{p{2.3cm}}{\raggedright \;\;\;\;\;\;\;\;\;\;\;Method} &
        \multicolumn{1}{p{2cm}}{\raggedleft Variation} &
        \multicolumn{1}{p{2cm}}{\centering AUROC(\%)} \\
        \midrule
        & CVDD  & best & 83.1 $\pm$ \small 4.4 \\
        
        & OCSVM & best & 84.0 $\pm$ \small 5.0 \\ 
        
        & ELECTRA & adapted for AD & 84.6  $\pm$ \small 4.5\\ %
        
        & \textbf{DATE} & \textbf{(Ours)} & \textbf{90.0} $\pm$ \small 4.2 \\ 

        \midrule
        \midrule
        A. & Anomaly score  & MP &  72.4 $\pm$ \small  3.7\\
        &         		    & NE &  73.1 $\pm$ \small  3.9\\
		\midrule
		B. &  Generator   & small & 89.3  $\pm$ \small 4.2 \\ 
		&                 & large   & 89.8  $\pm$ \small 4.4 \\
		\midrule
		C. & Loss function    & RTD only & 89.4 $\pm$ \small 4.4 \\
		&                 & RMD only & 85.9 $\pm$ \small 4.1 \\
        \midrule
		D. & Masking       & 5 masks  &  87.5 $\pm$ \small 4.5 \\
        & patterns         & 10 masks &  89.2 $\pm$ \small 4.3 \\
        &                  & 25 masks &  89.8 $\pm$ \small 4.3 \\
	    &                   & 100 masks & 89.8 $\pm$ \small 4.3 \\ 
        \midrule
        E. & Mask percent   & 15\% & 89.5 $\pm$ \small 4.1 \\ 
        &                   & 25\% & 89.5 $\pm$ \small 4.1 \\ 
		\bottomrule
    \end{tabular}
\end{center}
\caption{Ablation study. We show results for the competition and report ablation experiments which are only one change away from our best \textbf{DATE} configuration: \textbf{A}. $PL_{RTD}$; \textbf{B}. Rand \textbf{C}. RTD + RMD; \textbf{D}. 50 masks; \textbf{E}. 50\%. For the ELECTRA line, we use: A. $PL_{RTD}$; B. Rand; C. RTD only; D. Unlimited; E. 15\%.
A. The Anomaly Score used over classification probabilities shows that $PL_{RTD}$ (used in DATE) is the best in predicting anomalies, meaning that our self-supervised classification task is well defined, with few ambiguous samples; B. A learned Generator does not justify its training cost; C. RMD Loss proved to be complementary with RTD Loss, their combination (in DATE) increasing the score and stabilizes the training; D+E. (table from: \cite{DATE})}
\label{tab: ablation}
\end{table}

\noindent\textbf{A. Anomaly score.} We explore three anomaly scores introduced in the \emph{$E^{3}Outlier$} framework~\autocite{e3outlier} on semi-supervised and unsupervised AD tasks in Computer Vision: Maximum Probability (MP), Negative Entropy (NE), and our modified Pseudo Label ($PL_{RTD}$). These scores are computed using the softmax probabilities from the final classification layer of the discriminator. PL is an ideal score if the self-supervised task manages to build and learn well separated classes. The way we formulate our mask prediction task enables a very good class separation, as theoretically proved in detail in the supplementary material~\ref{apx:disjoint_patterns}. Therefore, $PL_{RTD}$ proves to be significantly better in detecting the anomalies compared with MP and NE metrics, which try to compensate for ambiguous samples.

\noindent\textbf{B. Generator performance.} We tested the importance of having a learned generator, by using a one-layer Transformer with hidden size 16 (small) or 64 (large). The \emph{random generator} proved to be better than both parameterized generators.

\noindent\textbf{C. Loss function.} For the final loss, we combined RTD (which sanctions the prediction per token) with our RMD (which enforces the detection of the mask applied on the entire sequence). We also train our model with RTD or RMD only, obtaining weaker results. This proves that combining losses with supervisions at different scales (locally: token-level and globally: sequence-level) improves AD performance. Moreover, when using only the RTD loss, the training can be very unstable (AUROC score peaks in the early stages, followed by a steep decrease). With the combined loss, the AUROC is only stationary or increases with time.

\noindent\textbf{D. Masking patterns.} The mask patterns are the root of our task formulation, hiding a part of the input tokens and asking the discriminator to classify them. As experimentally shown, having more mask patterns is better, encouraging increased expressiveness in the embeddings. Too many masks on the other hand can make the task too difficult for the discriminator and our ablation shows that having more masks does not add any benefit after a point. We validate the percentage of masked tokens in \textbf{E. Mask percent} ablation.

\subsection{Comparison with other AD methods}
\label{sec: ex-ssad}

\setlength{\tabcolsep}{4pt}
\begin{table}[t]
\begin{center}
	\begin{tabular}{l l r  r  r r  r }
		\toprule
		\multicolumn{1}{p{0.5cm}}{}&
        \multicolumn{1}{p{0.9cm}}{\raggedleft Inlier\\class} &
        \multicolumn{1}{p{1.8cm}}{\raggedleft iForest\\best} &
        \multicolumn{1}{p{1.8cm}}{\raggedleft OC-SVM\\best} &
        \multicolumn{1}{p{1.8cm}}{\raggedleft CVDD\\best} &
        \multicolumn{1}{p{1.8cm}}{\raggedleft mSVDD\\best} &
        \multicolumn{1}{p{1.8cm}}{\raggedleft \textbf{DATE (Ours)}} \\
        \midrule
        \parbox[t]{2mm}{\multirow{6}{*}{\rotatebox[origin=c]{90}{\textbf{20 News}}}} & 
        comp & 66.1 & 78.0 & 74.0  & 84.4 & \textbf{92.1}\\
        & rec & 59.4 & 70.0 & 60.6 & 77.7 &\textbf{83.4}\\
        & sci & 57.8 & 64.2 & 58.2 & 67.3 &\textbf{69.7}\\
        & misc & 62.4 &62.1& 75.7  & 76.2 &\textbf{86.0}\\
        &pol & 65.3 & 76.1 & 71.5  & 79.2 &\textbf{81.9}\\
        &rel & 71.4 & 78.9& 78.1   & 83.6 &\textbf{86.1}\\
        \midrule        
        \parbox[t]{2mm}{\multirow{4}{*}{\rotatebox[origin=c]{90}{\textbf{AG News}}}} & 
        business & 79.6& 79.9 & 84.0  & - & \textbf{90.0}\\
        & sci & 76.9 & 80.7 & 79.0    & - & \textbf{84.0}\\
        & sports &  84.7& 92.4 & 89.9 & - & \textbf{95.9}\\
        & world & 73.2 & 83.2 & 79.6  & - & \textbf{90.1}\\
		\bottomrule
    \end{tabular}
\end{center}
\caption{Semi-supervised performance (AUROC\%). We test on the 20Newsgroups and AG News datasets, by comparing DATE against several strong baselines and state-of-the-art solutions (with multiple variations, choosing the best score per split as detailed in Sec.~\ref{sec: ex-ssad}): iForest, OC-SVM, CVDD, and mSVDD. We largely outperform competitors with an average improvement of 4.7\% on 20Newsgroups and 6.9\% on AG News compared with the next best solution. Note that DATE uses the same set of hyper-parameters per dataset. (table adapted from: \cite{DATE})}
\label{tab: main_experiment}
\end{table}

\noindent
We compare our method against classical AD baselines like Isolation Forest \autocite{iso_forest} and existing state-of-the-art One Class SVM \autocite{ocsvm}, CVDD \autocite{acl2019}, and mSVDD \autocite{cvdd-2} We outperform all previously reported performances on all \emph{20Newsgroups} splits by a large margin: 13.5\% over the best reported CVDD and 11.7\% over the best OCSVM, and 4.7\% over the best mSVDD, as shown in Tab.~\ref{tab: main_experiment}. In contrast, DATE uses the same set of hyper-parameters for a dataset, for all splits. For a proper comparison, we keep the same experimental setup as the one introduced in \autocite{acl2019}.

\vspace{-1em}
\paragraph{Isolation Forest.} We apply it over fastText or GloVe embeddings, varying the number of estimators $(64, 100, 128, 256)$, and choosing the best model per split. In the unsupervised AD setup, we manually set the percent of outliers in the train set.

\vspace{-1em}
\paragraph{OCSVM.} We use the One-Class SVM model implemented in the CVDD work. For each split, we choose the best configuration (fastText vs Glove, rbf vs linear kernel, $\nu$ $\in$ [0.05, 0.1, 0.2, 0.5]).

\vspace{-1em}
\paragraph{CVDD.} For each split, we chose the best column out of all reported context sizes ($r$). The scores reported using the $c^*$ context vector depends on the ground truth and it only reveals ``\textit{the potential of contextual anomaly detection}'', as the authors mention.

\vspace{-1em}
\paragraph{mSVDD.} This model~\autocite{cvdd-2} is the current state-of-the-art solution for AD on text. For each split, we chose the best column out of all reported configurations, excluding those which use negative supervision.

\subsection{Unsupervised AD}
\label{sec: ex-uad}
\begin{figure}[b!]
\centering
\includegraphics[width=0.7\columnwidth]{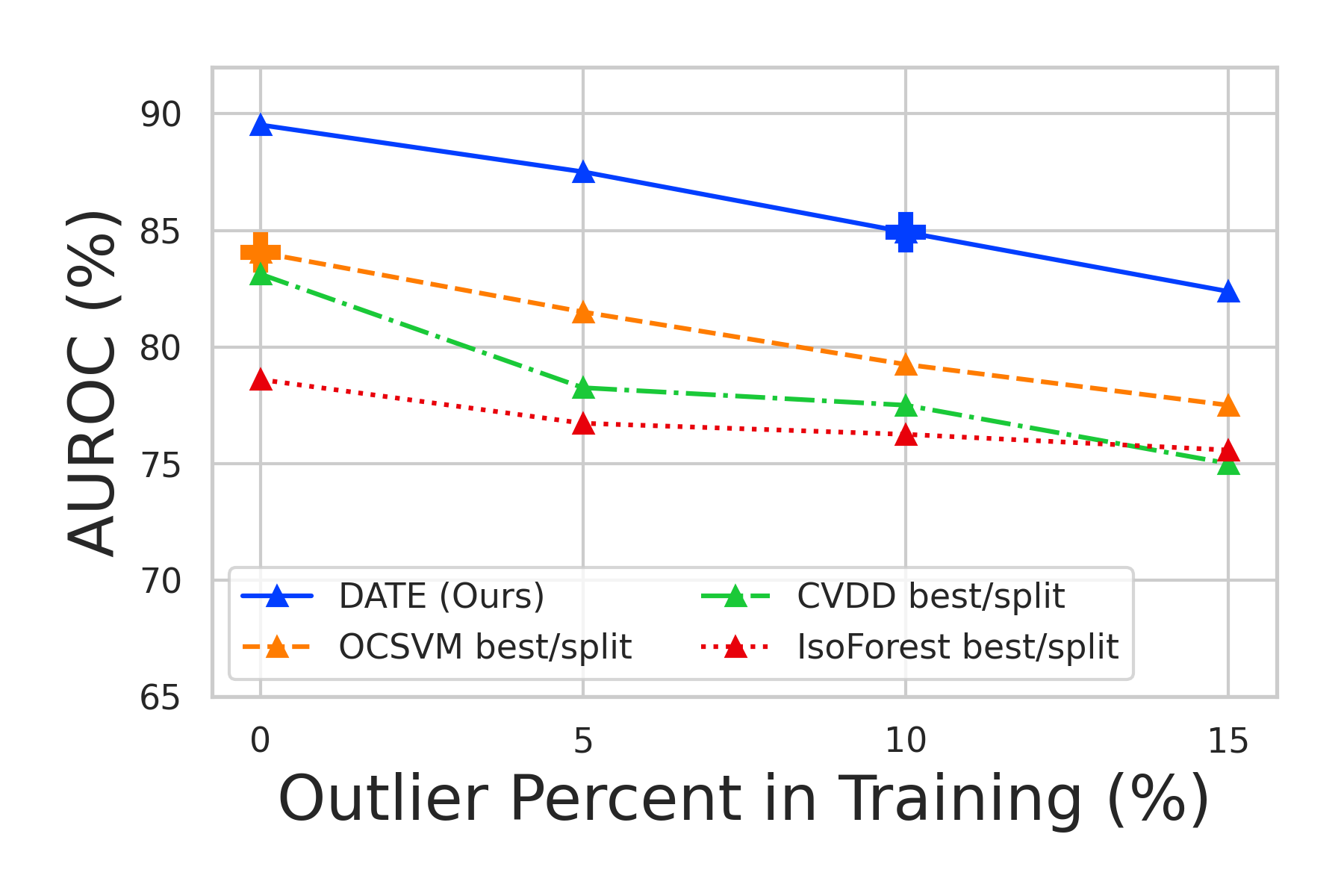}
\caption{Unsupervised AD. We test the performance of our method when training on impure data, which contains anomalies in various percentages: 0\%-15\%. The performance slowly decreases when we increase the anomaly percentage, but even at 10\% contamination, it is still better than state-of-the-art results on self-supervised anomaly detection in text \autocite{acl2019}, which trains on 0\% anomalous data, proving the robustness of our method. Experiments were done on all AG News splits. (figure from: \cite{DATE})}
\label{fig: fully_unsup_od}
\end{figure}

\noindent
We further analyse how our algorithm works in a fully unsupervised scenario, namely when the training set contains some anomalous samples (which we treat as normal ones). By definition, the quantity of anomalous events in the training set is significantly lower than the normal ones. In this experiment, we show how our algorithm performance is influenced by the percentage of anomalies in training data. Our method proves to be extremely robust, surpassing state-of-the-art, which is a semi-supervised solution, trained over a clean dataset (with 0\% anomalies), even at 10\% contamination, with +0.9\% in AUROC (see Fig.~\ref{fig: fully_unsup_od}). By achieving outstanding performance in the unsupervised setting, we make unsupervised AD in text competitive against other semi-supervised methods. The reported scores are the mean over all AG News splits. We compare against the same methods presented in Sec.~\ref{sec: ex-ssad}.

\subsection{Qualitative results}
\begin{figure}
    \centering
    \hspace{-12em}
    \begin{minipage}{.49\linewidth}
            \begin{subfigure}[t]{1.2\linewidth}
                \includegraphics[width=0.49\columnwidth]{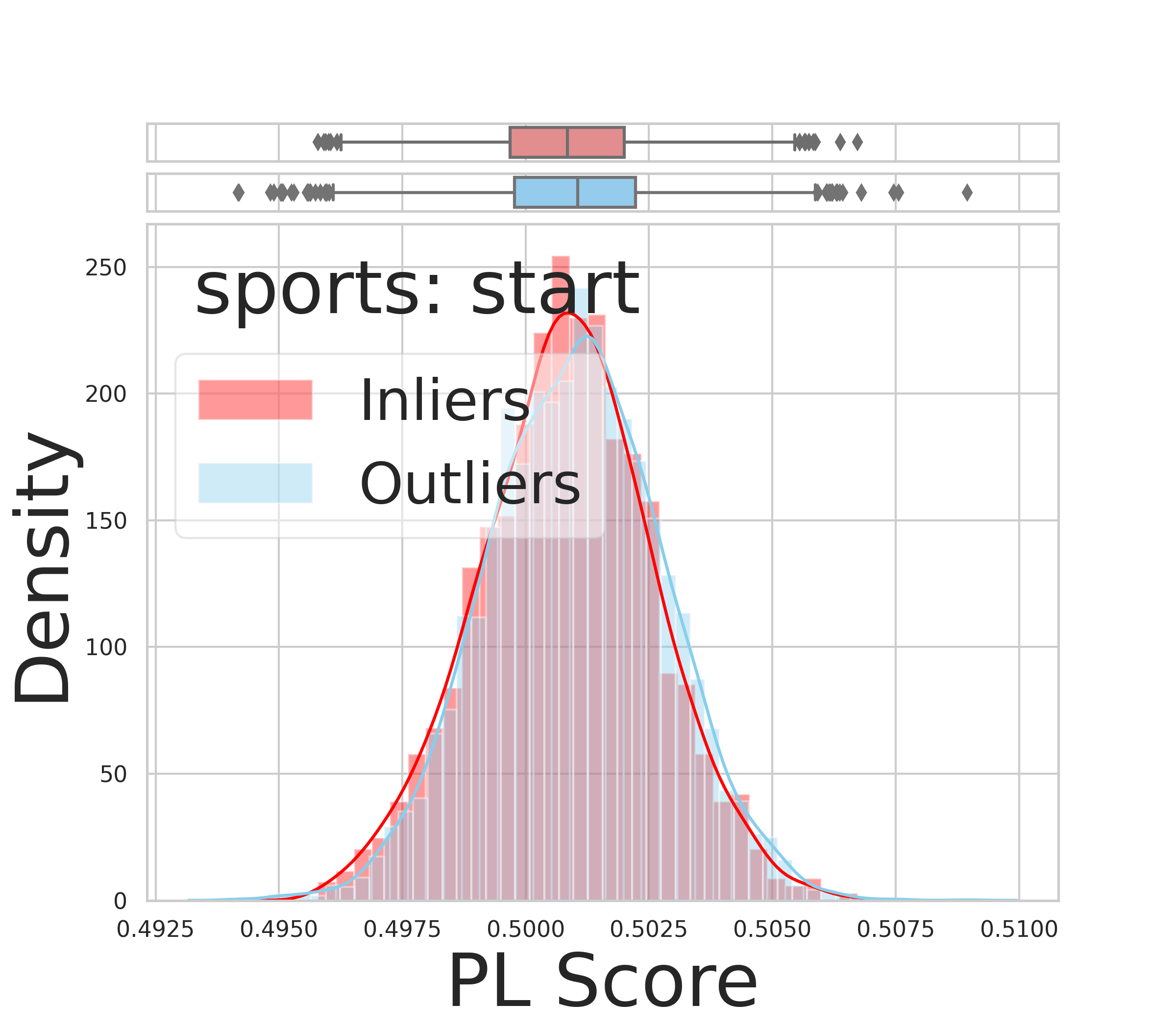}
                \includegraphics[width=0.49\columnwidth]{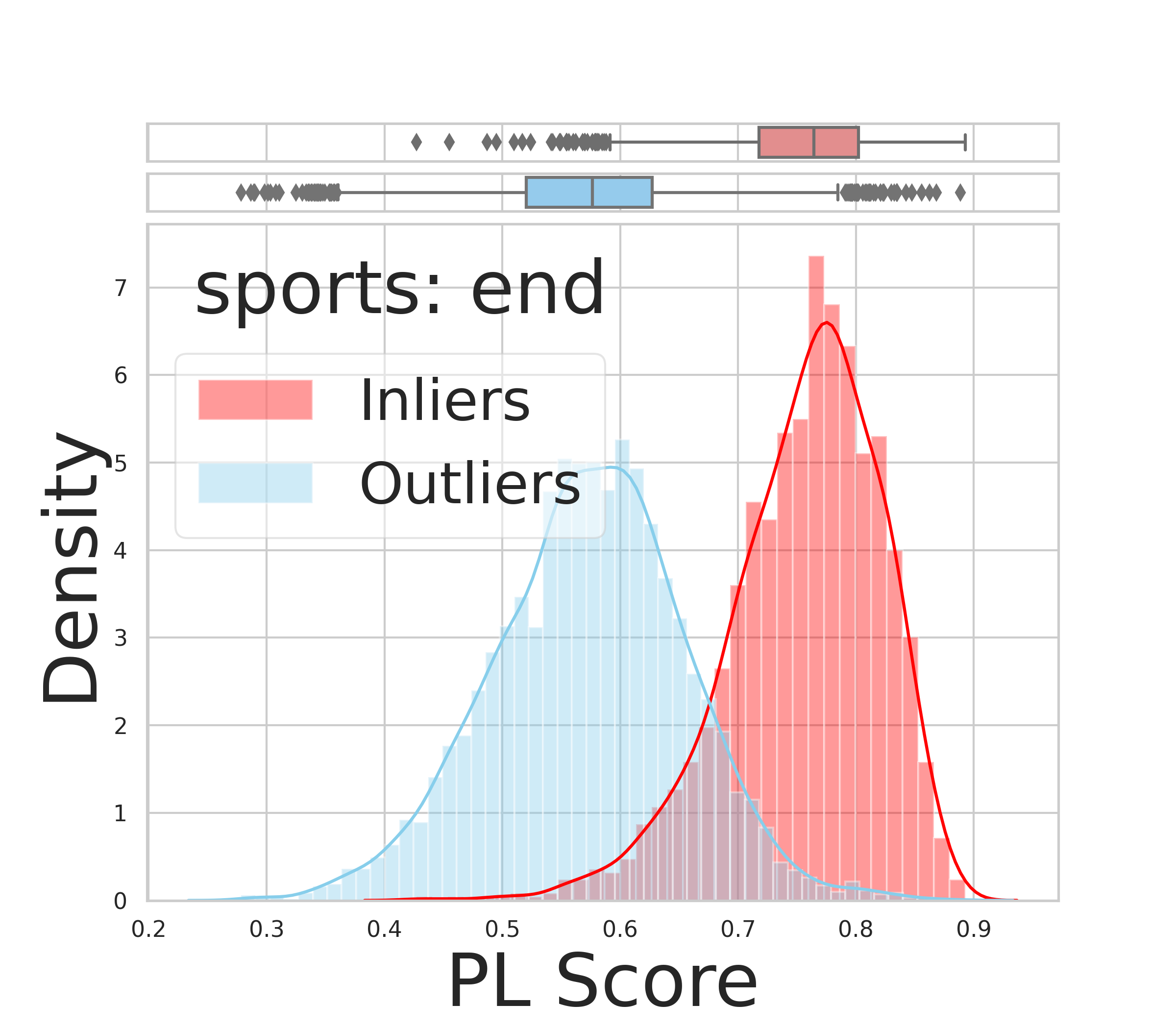} 
                \includegraphics[width=0.49\columnwidth]{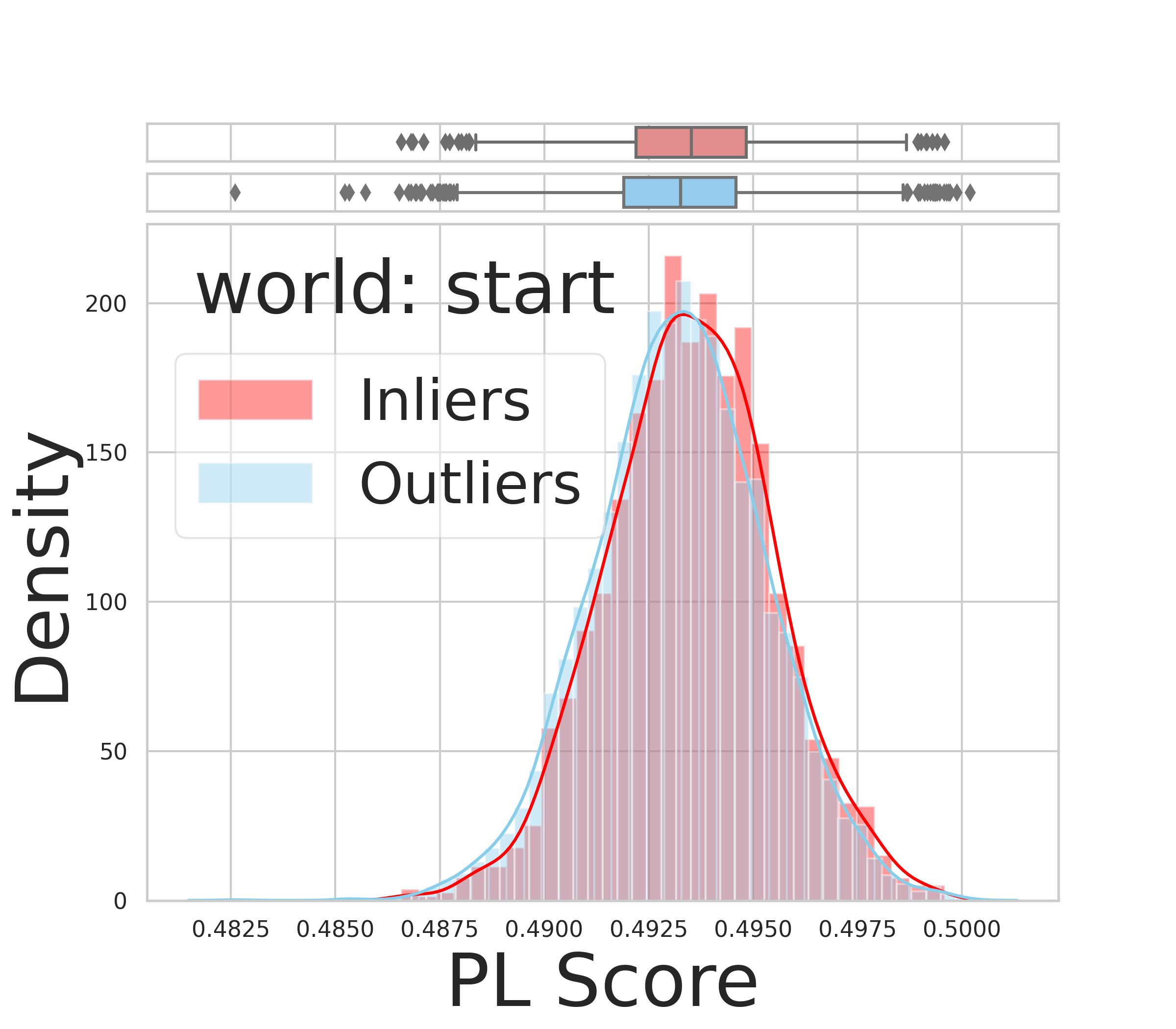}
                \includegraphics[width=0.49\columnwidth]{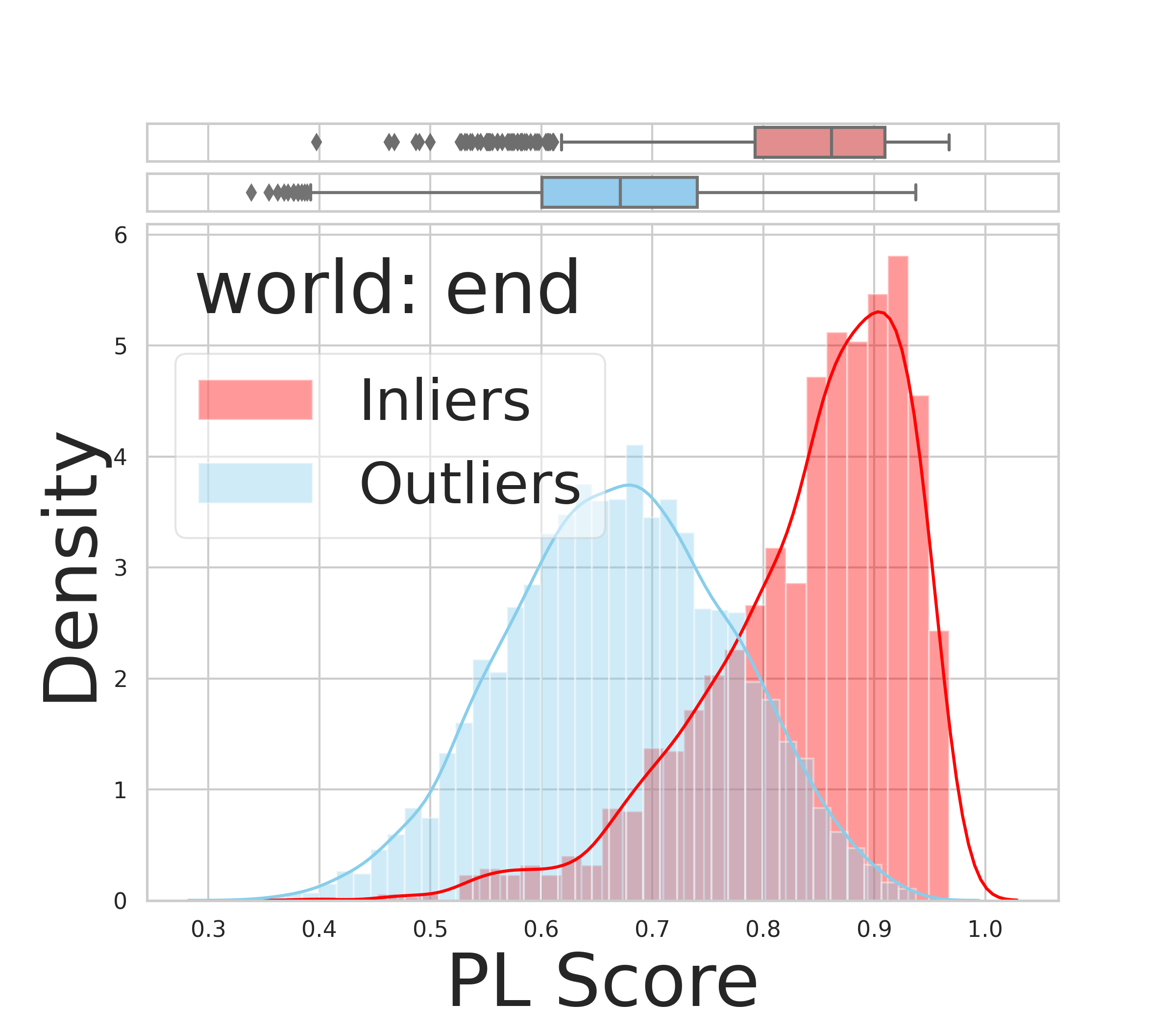}
            \caption{Normalized histogram for anomaly score. We see how the anomaly score (PL) distribution varies among inliers and outliers, from the beginning of the training ($1^{st}$ column) to the end ($2^{nd}$ column), where the two become well separated, with relatively low interference between classes. Note that a better separation is correlated with high performance ($1^{st}$ line split has $95.9\%$ AUROC, while the $2^{nd}$ has only $90.1\%$).}
            \label{fig: histogram}

            \end{subfigure}
        \end{minipage}
        \hspace{5em}
    \begin{minipage}{.49\linewidth}
        \begin{subfigure}[b]{1.4\linewidth}
            \includegraphics[width=0.99\columnwidth]{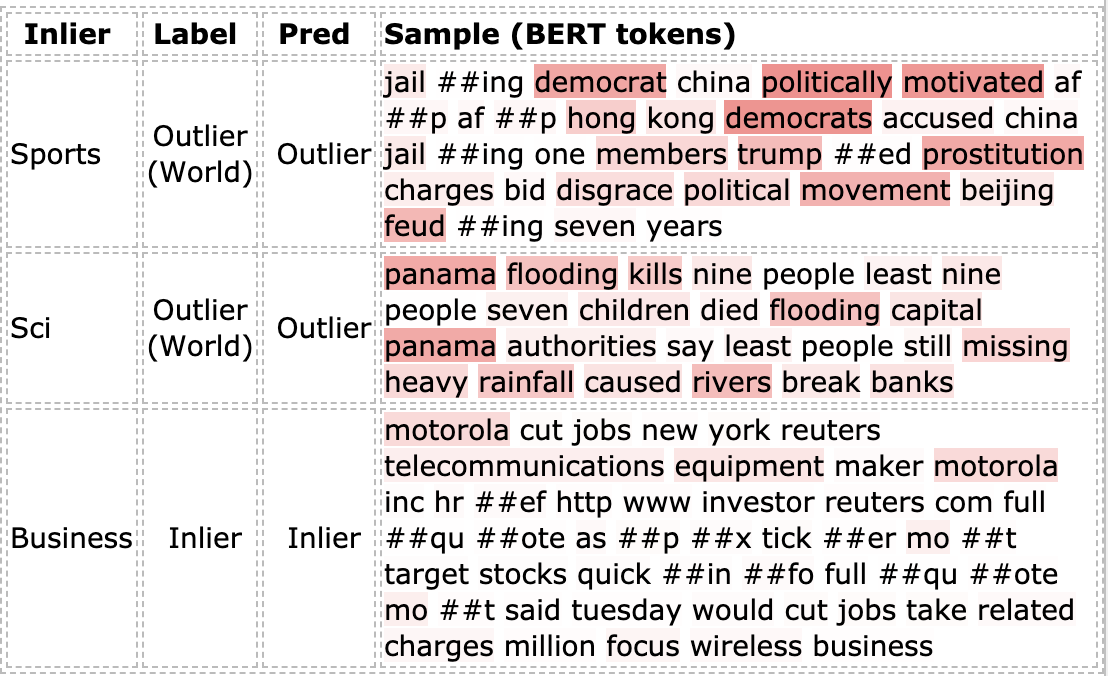} 
            \caption{Qualitative examples. Lower scores are shown in a more intense red, and point to anomalies. In the $1^{st}$ example, words from politics are flagged as anomalous for sports. In the $2^{nd}$ one, words describing natural events are outliers for technology. In the $3^{rd}$ row, while few words have higher anomaly potential for the business domain, most of them are appropriate.}
            \label{fig: qual_ex}
        \end{subfigure}
    \end{minipage}
    \caption{Qualitative results. a) Evolution of the anomaly scores on the sports and sci test sets. b) Anomalous words for three different scenarios. (figures from: \cite{DATE})}
\end{figure}

We show in Fig.~\ref{fig: qual_ex} how DATE performs in identifying anomalies in several examples. Each token is colored based on its PL score.

We also show how our anomaly score (PL) is distributed among normal vs abnormal samples. For visualization, we chose two splits from AG News and report the scores from the beginning of the training to the end. We see in Fig.~\ref{fig: histogram} that, even though at the beginning, the outliers' distribution of scores fully overlaps with the inliers, at the end of training the two are well separated, proving the effectiveness of our method.



\chapter{Conclusion and Further Developments}
\label{chp:conclusion}

\noindent
We propose DATE, a model for tackling Anomaly Detection in Text, and formulate an innovative self-supervised task, based on masking parts of the initial input and predicting which mask pattern was used. After masking, a generator reconstructs the initially masked tokens and the discriminator predicts which mask was used.
We optimize a loss composed of both token and sequence-level parts, taking advantage of powerful supervision, coming from two independent pathways, which stabilizes learning and improves AD performance.
For computing the anomaly score, we alleviate the burden of aggregating predictions from multiple transformations by introducing an efficient variant of the Pseudo Label score, which is applied per token, only on the original input.
We show that this score separates very well the abnormal entries from normal ones, leading DATE to outperform state-of-the-art results on all AD splits from 20Newsgroups and AG News datasets, by a large margin, both in the semi-supervised and unsupervised AD settings.

We would like to further investigate and analyse other self-supervision tasks for AD in text. In its current form, DATE is randomly masking tokens from the input sequence. It would be interesting to see how masking specific parts of speech would affect the anomaly detection capabilities. Another intriguing approach to self-supervision is by contrastive learning \autocite{simclr1}. One way in which we could adapt DATE to contrastive learning is to apply different augmentations (by applying different masking patterns) and minimize the cosine distance between the representations which come from the same text corrupted in different ways while maximizing the cosine distance between different texts. An alternative way of generating anomaly scores is by using energy-based models. Recently, EBMs have become increasingly popular. An EBM can theoretically be used as an anomaly detector by directly evaluating the energy function \autocite{anomaly-review} to produce anomaly scores. Models such as ELECTRA can be formulated as energy-based models \autocite{electric}, this raises the question if they can also be used to detect anomalies. Finally, DATE demonstrates capabilities for authorship detection and can detect stylistic differences between texts written by different authors (see supplementary material \ref{apx:authorship}). We would like to further evaluate deep anomaly detectors for such tasks.

\clearpage

\printbibliography

\begin{appendices}
\chapter{Hyperparameters and implementation}

\section{Disjoint patterns analysis}
\label{apx:disjoint_patterns}
\noindent
We start from two observations regarding the performance of DATE, our Anomaly Detection algorithm. First, a discriminative task performs better if the classes are well separated \autocite{Deng2012} and there is a low probability for confusions. Second, the PL score for anomalies achieves the best performance when the probability distribution for its input is clearly separated. Intuitively, for three classes, PL([0.9, 0.05, 0.05]) is better than PL([0.5, 0.3, 0.2]) because it allows PL to give either near $1$ score if the class is correct, either near $0$ score if it is not, avoiding the zone in the middle where we depend on a well chosen threshold.

Since the separation between the mask patterns greatly influences our final performance, we next analyze our AD task from the mask pattern generation point of view. Ideally, we want to have a sense of how disjoint our randomly sampled patterns are and make an informed choice for the pattern generation hyper-parameters.

First, we start by computing an upper bound for the probability of having two patterns with at least $p$ common masked points. We have $\binom{S}{M}$ patterns, where $S$ is the sequence length and $M$ is the number of masked tokens. We fix the first $p$ positions that we want to mask in any pattern. Considering those fixed masks, the probability of having a sequence with $M$ masked tokens, with $p$ tokens in the first positions is $r$:

\begin{equation}
\label{eq: r_ratio}
    r = \frac{\binom{S-p}{M-p}}{\binom{S}{M}}.
\end{equation}

Next, the probability that two sequences mask the first $p$ tokens is $r^2$. But we can choose those two positions in a $\binom{S}{p}$ ways. So the probability that any two sequences have at least $p$ common masked tokens is lower than $UB_{2}$:

\begin{equation}
\label{eq: concide_bound}
    UB_{2} = \binom{S}{p} r^2\\
\end{equation}

Next, out of our generated patterns, we sample $N$ masks, so the probability becomes less than the upper bound $UB_{N}$:
\begin{equation}
\label{eq: final_bound}
    \begin{aligned}
    UB_{N} &= \binom{N}{2} UB_{2} = \binom{N}{2} \binom{S}{p} r^2 \\ 
        &= \binom{N}{2} \binom{S}{p} (\frac{\binom{S-p}{M-p}}{\binom{S}{M}})^2.
    \end{aligned}
\end{equation}

In our experiments, the sequence length is $S = 128$ and we chose the number of masked tokens to be between 15\% and 50\% ($M$ between 19 and 64). We consider that two patterns are disjoint when they have less than $p$ masked tokens in common, for $N$ sampled patterns.

The probability that any two patterns collide (have more than $p$ masked tokens in common) is very low. We compute several values for its upper bound: $UB_{N=100, p=12}=5e-4$, $UB_{N=100, p=15}=1e-9$, $UB_{N=10, p=15}=1e-11$, $UB_{N=10, p=13}=1e-7$.

In conclusion, for our specific setup, the probability for two masks to largely overlap (large $p$ compared with $S$) is extremely small, ensuring us a good performance in the discriminator. We take advantage of this property of our pretext task by combining the discriminator output probabilities with the PL score.

\section{Model implementation}
\label{apx:model_implementation_details}
\noindent
We add next more details on the implementation of the modules: from the ablation experiments in Tab.~\ref{tab: ablation}, \textbf{Generator (small)}: 1 Transformer layer, with 4 self-attention heads, token and positional embeddings of size 128, hidden layer of size 16, feedforward layer of sizes 1024 and 16; \textbf{Generator (large)}: 1 Transformer layer, with 4 self-attention heads, token and positional embeddings of size 128, hidden layer of size 64, feedforward layer of sizes 1024 and 64;
As empirical experiments showed us, we choose a \emph{random Generator} (samples were drawn from a uniform distribution over the vocabulary) in our final model.
\textbf{Discriminator}: 4 Transformer layers, each with 4 self-attention heads, hidden layers of size 256, feedforward layers of sizes of 1024 and 256, 128-dimensional token and positional embeddings, which are \emph{tied} with the generator. For other unspecified hyper-parameters we use the ones in ELECTRA-Small model.
\textbf{Prediction Heads}: both heads have 2 linear layers separated by a non-linearity, ending in a classification. \textbf{Loss weights:} We set the RTD $\lambda$ weight to 50 as in \autocite{Clark2020}, and the RMD $\mu$ weight to 100.

\chapter{Other results}

\section{More qualitative and quantitative results on AG News}
\label{apx:prauc_experiments}
\begin{figure}[!b]
    \centering
    \includegraphics[width=0.8\textwidth]{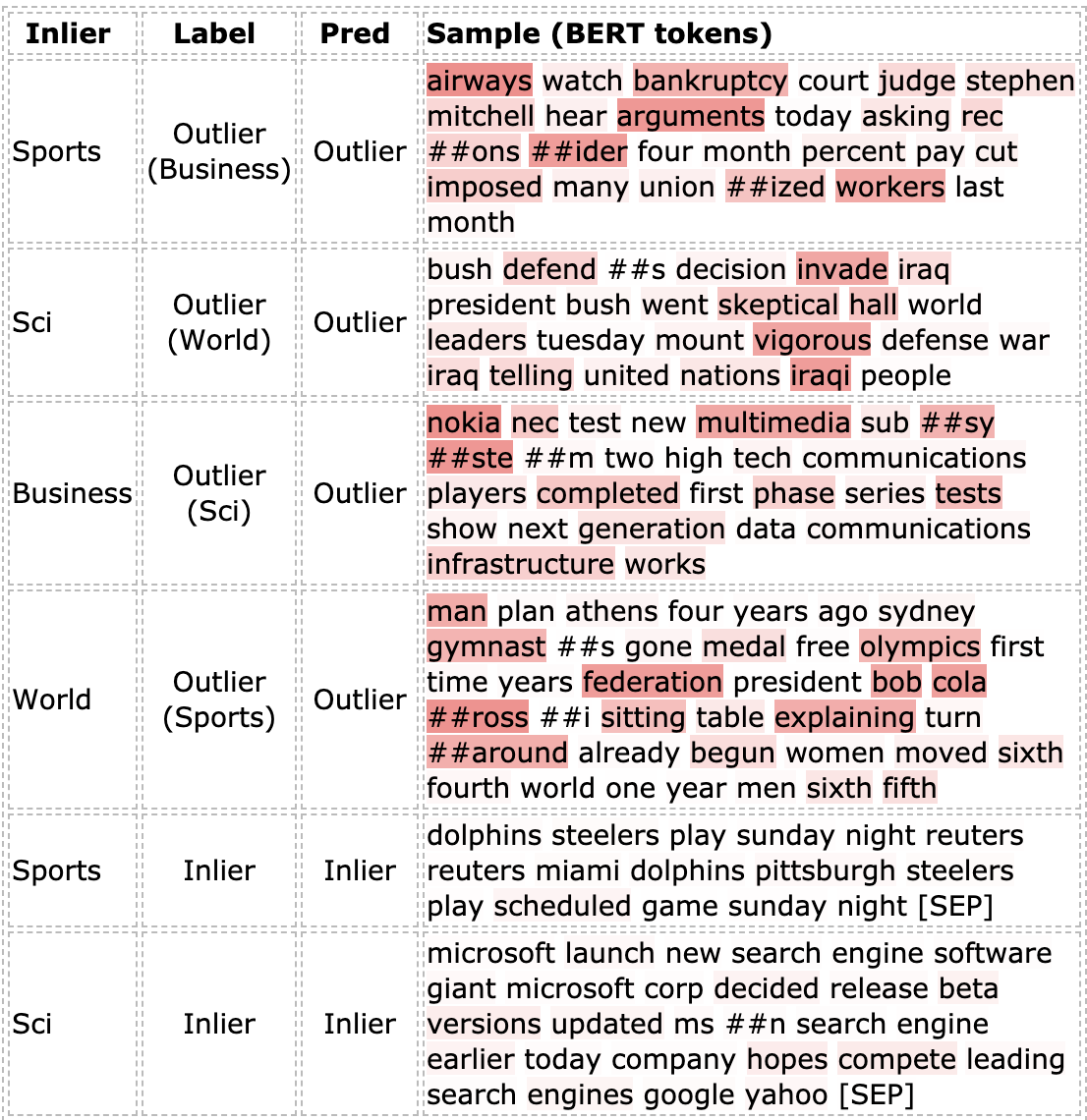} 
    \caption{More qualitative examples.}
    \label{fig: qual_ex_2}
\end{figure}

\noindent
In Fig.~\ref{fig: qual_ex_2} we show more qualitative results, trained on different inliers. To encourage further more detailed comparisons, we report the AUPR metric on AG News for inliers and outliers (see Tab.~\ref{tab:prauc_experiments}). When all the other metrics are almost saturated, we notice that AUPR-in better captures the performance on a certain split.


\vspace{-9em}

\begin{table}[!b]
\centering
\begin{tabular}{l  r r r r}
\toprule
\multicolumn{1}{p{1cm}}{Subset} & business & sci & sports & world\\
\midrule
AUPR-in  & 74.8 & 62.4 & 88.8 & 81.9 \\
AUPR-out & 96.1 & 93.5 & 98.5 & 95.5 \\

\bottomrule
\end{tabular}
\caption{We report AUPR metric for AG News splits, on inliers and outliers since this is a more relevant metric for unbalanced classes (which is the case for all splits in text AD, as explained in Anomalies setup).}
\label{tab:prauc_experiments}
\end{table}

\clearpage

\section{Authorship detection}
\label{apx:authorship}

\begin{table}[b!]
\begin{center}
    \hspace{-4.3em}
	\begin{tabular}{l| r | r | r | r | r | r | r | r | r | r | r }
		\toprule
        \multicolumn{1}{p{2cm}}{\raggedleft train/val} &
        \multicolumn{1}{p{0.8cm}}{\raggedleft Usr1} & 
        \multicolumn{1}{p{0.8cm}}{\raggedleft Usr2} & 
        \multicolumn{1}{p{0.8cm}}{\raggedleft Usr3} & 
        \multicolumn{1}{p{0.8cm}}{\raggedleft Usr4} & 
        \multicolumn{1}{p{0.8cm}}{\raggedleft Usr5} & 
        \multicolumn{1}{p{0.8cm}}{\raggedleft Usr6} & 
        \multicolumn{1}{p{0.8cm}}{\raggedleft Usr7} & 
        \multicolumn{1}{p{0.8cm}}{\raggedleft Usr8} & 
        \multicolumn{1}{p{0.8cm}}{\raggedleft Usr9} & 
        \multicolumn{1}{p{0.8cm}}{\raggedleft Usr10} \\ 

        \midrule
        25/5 (AUROC)    & 100 & 100 & 98.2& 99.6 & 100 & 100 & 96.9 & 100 & 98.4 & 100 \\
        25/5 (AUPR-in)  & 100 & 100 & 87.2& 93.8 & 100 & 100 & 65.7 & 100 & 84.1 & 100 \\
        25/5 (AUPR-out) & 100 & 100 & 99.9& 99.9 & 100 & 100 & 99.8 & 100 & 99.9 & 100 \\
        \midrule
        15/15 (AUROC)    & 96.7 & 98.5 & 97.3& 99.3 & 98.2 & 100 & 95.2 & 100 & 93.8 & 98.6 \\
        15/15 (AUPR-in)  & 68.2 & 83.6 & 74.6& 89.9 & 85.0 & 100 & 57.5 & 100 & 61.1 & 89.9 \\
        15/15 (AUPR-out) & 99.8 & 64.4 & 99.9& 99.9 & 99.9 & 100 & 99.7 & 100 & 99.6 & 99.9 \\
        \midrule
        5/25 (AUROC)    & 82.7 & 89.1 & 81.7& 88.6 & 83.2 & 97.4 & 79.4 & 82.3 & 67.7 & 84.4 \\
        5/25 (AUPR-in)  & 42.3 & 64.4 & 39.1& 52.8 & 56.1 & 83.9 & 34.6 & 65.9 & 22.0 & 42.0 \\
        5/25 (AUPR-out) & 98.8 & 89.1 & 98.7& 99.3 & 98.7 & 99.9 & 98.3 & 97.8 & 97.3 & 99.0 \\
        \midrule

        1/29 (AUROC) & 72.7 & 82.0 & 73.2    & 87.2 & 82.2 & 95.4 & 83.2 & 80.3 & 53.1 & 65.1 \\
        1/29 (AUPR-in) & 20.2 & 33.1 & 15.3  & 43.6 & 41.2 & 66.8 & 29.6 & 42.0 & 12.3 & 17.8 \\
        1/29 (AUPR-out) & 98.1 & 98.8 & 97.7 & 99.2 & 98.7 & 99.7 & 98.9 & 97.9 & 95.5 & 97.2 \\
        \midrule
		\bottomrule
    \end{tabular}
\end{center}
\caption{Semi-supervised AD Performance (AUROC\%, AUPR-in\%, AUPR-out\%) on the PAN2020 dataset (custom setup) using DATE: The train/val ratio is the number of samples for training per author, and number of outlier articles per author when testing respectively. No hyperparameter tuning was done. The AUPR-out scores are very high because AUPR is sensible to imbalanced data.}
\label{tab: pan_exp}
\end{table}

\begin{figure}[t!]
    \centering
    \begin{subfigure}[b]{1.3\textwidth}
        \centering
        \includegraphics[width=0.475\linewidth]{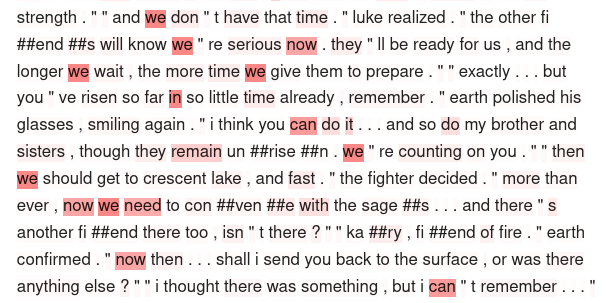}%
        \hfill
        \includegraphics[width=0.475\linewidth]{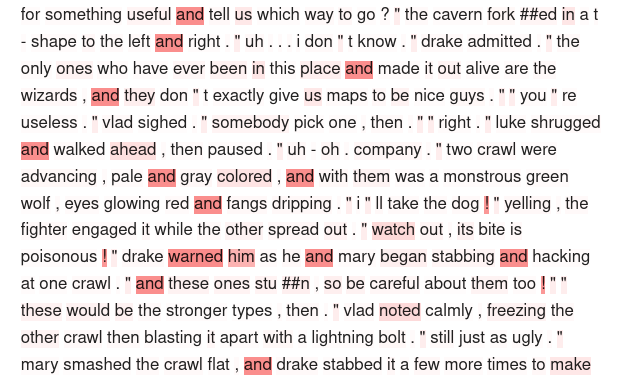}
        \caption{\textit{Usr9} correctly identified as \textit{Usr9}. Interestingly, the network identifies the use of stop-words as being anomalous.}
    \end{subfigure}
    \vskip\baselineskip
    \begin{subfigure}[b]{1.3\textwidth}
        \centering
        \includegraphics[width=0.475\linewidth]{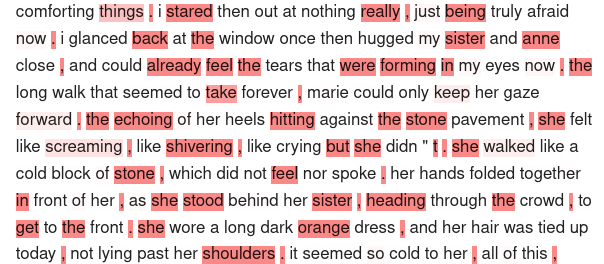}%
        \hfill
        \includegraphics[width=0.475\linewidth]{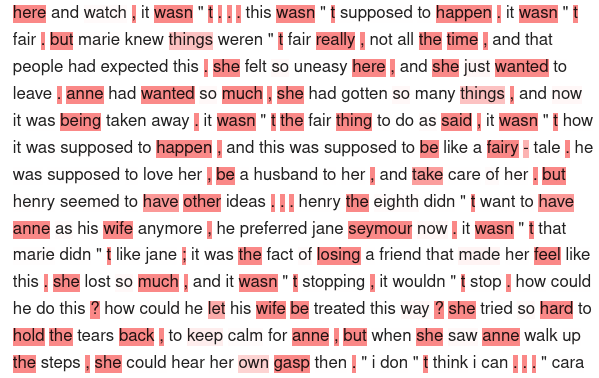}
        \caption{Network trained on \textit{Usr9}. The network correctly classifies the texts as not being written by \textit{Usr9}. Notice the fact that the network considers that the use of punctuation is dubious.}
    \end{subfigure}
    \caption{DATE Qualitative results on the PAN2020 dataset. Lower PL scores are shown in a more intense red and contribute more to a higher anomaly probability.}
    \label{qual:pan}
\end{figure}

The PAN 2020 \autocite{pan20} data is collected from the FanFiction website\sidenote{\url{https://www.fanfiction.net/}} and represents an excerpt of one of the four shared tasks on authorship analysis presented at PAN 2020, namely, the second one. The data is aimed at cross-domain authorship verification, with the purpose of understanding the associations between authors and their texts in a context agnostic to domain-specific vocabulary. 

We've sampled the top 19 authors from our curated dataset, each of them having written 30 articles. We've then picked a subset of 10 authors from the initial 19 for training and testing (due to time and resource constraints). We're treating the Authorship Verification task as a semi-supervised Anomaly Detection problem in the following way - Let $A_1, ..., A_{19}$ be the texts written by the authors, and $\mathcal{A} = \{A_1, ..., A_{19}\}$ be the set of all texts, then given an $A_k \in \mathcal{A}$, we set $A_k$ to be the set of normal texts, and $\mathcal{A} \setminus {A_k}$ to be the set of anomalous texts.

We are trying to profile authors using DATE. We are using the same hyperparameters as in the ``AG News'' dataset experiments. The only preprocessing that was done to the data is making it lower-cased. We're reporting scores at 1500 steps into the training process. The results of the experiments are in Table \ref{tab: pan_exp}.

Figure \ref{qual:pan} shows how DATE is able to detect stylistic changes when correctly profiling an author. The model considers that certain punctuation and word usage is dubious. When scoring an inlier entry, DATE seems to think that some stopword usage is anomalous, this could indicate a change of stylistic approach when an author is writing texts in different fandoms.

\section{Music genre detection}
\label{apx:music}
\noindent
The Song Lyrics dataset\sidenote{Many thanks to Matei Bejan for collecting this dataset: \href{https://www.kaggle.com/mateibejan/multilingual-lyrics-for-genre-classification}{Kaggle multilingual lyrics for genre classification}} is composed of four sources and consists of over 290,000 multilingual song lyrics and their respective genres: Rock, Metal, Pop, Hip-Hop, Electronic, R\&B, Country, Folk and Indie. The initial data was forwarded from the 2018 Textract Hackathon\sidenote{\url{https://www.sparktech.ro/textract-2018/}}. This was enhanced with data from three kaggle datasets: \textit{150K Lyrics Labeled with Spotify Valence}, \textit{dataset lyrics musics} and \textit{AZLyrics song lyrics}.

Apart from the original 2018 Textract data, the other datasets were not provided with a Genre feature. In order to deal with the lack of labels, a labeling function was built using the spotipy library, which uses the Spotify API in order to retrieve the genre of an Artist. The Spotify API returns a list of genres for one artist, so only the most common genre to be said artists dominant genre was considered. The results are shown in Tables \ref{tab: quant_newsmusic} and \ref{tab: cvdd_hiphop}.

\begin{table}[t!]
\begin{center}
    \caption{Semi-supervised AD Performance (AUC-ROC\%) on the Music Genres datasets. We compare two classical models (Isolation Forest, One-Class SVM) with two neural mdoels (CVDD, DATE). We are choosing the best score per split, as detailed in Sec. \ref{sec: ex-ssad}. DATE is using the same hyperparameters per dataset. Interestingly, every model is able to outperform DATE on the Hip-Hop subset, CVDD having the largest margin (+24.1\% ROC-AUC).}
    \label{tab: quant_newsmusic}
	\begin{tabular}{l l| r | r | r | r |}
		\toprule
		\multicolumn{1}{p{0.2cm}}{}&
        \multicolumn{1}{p{2cm}}{\raggedleft Inlier class} &
        \multicolumn{1}{p{2cm}}{\raggedleft iForest} &
        \multicolumn{1}{p{2cm}}{\raggedleft OC-SVM} &
        \multicolumn{1}{p{2cm}}{\raggedleft CVDD} &
        \multicolumn{1}{p{2cm}}{\raggedleft DATE} \\
        \midrule
        \parbox[t]{5mm}{\multirow{10}{*}{\rotatebox[origin=c]{90}{\textbf{Music Genres}}}} &
        Rock & 45.6 & 44.2 & 45.1 & \textbf{54.4}\\
        & Electronic & 48.5 & 48.8 & 45.4 & \textbf{56.0}\\
        & Country & 57.3 & 57.5 & 54.3 & \textbf{70.8}\\
        & Metal & 55.9 & 46.6 & 48.5 & \textbf{56.6}\\
        & Indie & 51.8 & 47.1 & 45.5 & \textbf{57.8}\\            
        & Jazz & 53.5 & 52.0 & 56.8 & \textbf{70.1}\\
        & Folk & 49.5 & 49.5 & 47.0 & \textbf{53.4}\\
        & R\&B & 58.4 & 58.1 & 61.3 & \textbf{71.0}\\
        & Hip-Hop & 65.0 & 72.9 & \textbf{84.3} & 60.2\\
        & Pop & 43.2 & 43.7 & 61.3 & \textbf{62.9}\\
        \midrule
        & Mean & 52.9 & 52.0 & 55.0 & \textbf{61.3}\\
        \bottomrule
    \end{tabular}
    \end{center}
\end{table}

\begin{table}[b!]
\small
\begin{center}
	\begin{tabular}{c | c | c | c | c}
		\toprule
        \multicolumn{1}{p{1.8cm}}{\centering Context 1} &
        \multicolumn{1}{p{1.8cm}}{\centering \textbf{Context 2}} &
        \multicolumn{1}{p{1.8cm}}{\centering Context 3} &
        \multicolumn{1}{p{1.8cm}}{\centering Context 4} &
        \multicolumn{1}{p{1.8cm}}{\centering Context 5} \\

        \midrule
        'm & \textbf{livin} & to & the & que\\
        \midrule
        i & \textbf{turnin} & can & of & mas \\
        \midrule
        're & \textbf{shakin} & could & in & como\\
        \midrule
        y' & \textbf{waitin} & make & 's & porque\\
        \midrule
        somebody & \textbf{walkin} & will & that & sabe\\
        \midrule
        myself & \textbf{keepin} & would & is & desde\\
        \midrule
        everybody & \textbf{slippin} & pray & this & boca\\
		\bottomrule
    \end{tabular}
\end{center}
\caption{CVDD context vectors. The second context is the most meaningful one when detecting anomalies. Notice that the second context contains the colloquial forms of verbs. The first context seems to cluster pronouns together. The third and forth contexts contains modal verbs and stopwords. The last context vector is clustering together words in Spanish.}
\label{tab: cvdd_hiphop}
\end{table}

\clearpage

\section{Native language detection}
\label{apx:native_lang}
\noindent
ENNTT \autocite{Nisioi2016ACO} is an English corpus of original and human-translated texts extracted from the discussions that took place inside the European Parliament. The dataset consists of three types of texts, generated by either native English speakers, non-native English speakers or translated from another European language.
It contains 116,341 native sentences, 29,734 non-native sentences and 738,597 translated sentences.

The results when using both CVDD and DATE for one-class classification on ENNTT are poor, as can be seen in Table \ref{tab:enntt}.

    \begin{table}[t!]
    \begin{center}
    	\begin{tabular}{l| r | r |}
    		\toprule
            \multicolumn{1}{p{1cm}}{\raggedleft Outlier} &
            \multicolumn{1}{p{1.5cm}}{\raggedleft CVDD} &
            \multicolumn{1}{p{1.5cm}}{\raggedleft DATE} \\
            \midrule
            Translations & 43.2 & \textbf{55.4}\\
            Non-native & 52.3 & \textbf{56.0}\\
            \bottomrule
        \end{tabular}
    \end{center}
    \caption{Semi-supervised AD Performance (AUROC\%) on the ENNTT dataset. The performance is weak on both CVDD and DATE.}
    \label{tab:enntt}
    \end{table}

\end{appendices}

\end{document}